%% file: main.tex
\pdfoutput=1

\documentclass[11pt]{article}

\usepackage[]{acl}

\usepackage{txfonts}
\usepackage{latexsym}
\usepackage{multirow}
\usepackage{color,soul}
\usepackage{booktabs}
\usepackage[frozencache=true,cachedir=minted-cache]{minted}
\usepackage{amssymb}
\usepackage{longtable}
\usepackage{graphicx}

\usepackage{subcaption}

\usepackage[utf8]{inputenc}

\usepackage{microtype}

\usepackage{inconsolata}

\usepackage{pifont}

\usepackage{array}
\newcommand{\PreserveBackslash}[1]{\let\temp=\\#1\let\\=\temp}
\newcolumntype{C}[1]{>{\PreserveBackslash\centering}p{#1}}
\newcolumntype{R}[1]{>{\PreserveBackslash\raggedleft}p{#1}}
\newcolumntype{L}[1]{>{\PreserveBackslash\raggedright}p{#1}}

\definecolor{green}{RGB}{0, 128, 1}
\definecolor{orange}{RGB}{255, 90, 0}
\colorlet{lightorange}{orange!40}
\colorlet{lightyellow}{yellow!40}
\sethlcolor{lightorange}

\usepackage{listings}
\lstdefinelanguage{gpt}{
    morekeywords={},
    otherkeywords={<|im_start|>system, <|im_start|>user, <|im_start|>assistant, <|im_end|>, <<SYS>>, [INST], <</SYS>>, [/INST], Użytkownik:, Asystent:, <s>},
    sensitive=false,
    keywordstyle=\bfseries\color{green},
}

\usepackage[T1]{fontenc}

\usepackage[utf8]{inputenc}

\usepackage{microtype}

\newcommand*\rot{\rotatebox{90}}
\newcommand{\down}[1]{\raisebox{-.4ex}{\scriptsize #1}}

\title{Evaluation of Few-Shot Learning for Classification Tasks in the Polish Language}

\author{Tsimur Hadeliya,
        Dariusz Kajtoch\\
        ML Research at Allegro, Poznań, Poland \\
        \texttt{\{firstname.lastname\}@allegro.com}}

\begin{document}
\maketitle
\begin{abstract}
We introduce a few-shot benchmark consisting of 7 different classification tasks native to the Polish language. We conducted an empirical comparison with 0 and 16 shots between fine-tuning, linear probing, SetFit, and in-context learning (ICL) using various pre-trained commercial and open-source models. Our findings reveal that ICL achieves the best performance, with commercial models like GPT-3.5 and GPT-4 attaining the best performance. However, there remains a significant 14 percentage points gap between our best few-shot learning score and the performance of HerBERT-large fine-tuned on the entire training dataset.
Among the techniques, SetFit emerges as the second-best approach, closely followed by linear probing. We observed the worst and most unstable performance with non-linear head fine-tuning. Results for ICL indicate that continual pre-training of models like Mistral-7b or Llama-2-13b on Polish corpora is beneficial. This is confirmed by the improved performances of Bielik-7b and Trurl-13b, respectively. To further support experiments in few-shot learning for Polish, we are releasing handcrafted templates for the ICL.
\end{abstract}

\section{Introduction}
Few-shot learning enables models to grasp concepts from a limited number of examples~\citep{tunstall2022efficient, ye-etal-2021-crossfit, kwon2023efficient}, reducing the resources needed for data curation and training. This efficiency lowers the barriers to introducing new tasks, allowing for quicker deployment and more exploration. Over time, researchers have developed new methods for learning from a few examples. Progress has been driven mainly by fine-tuning pre-trained transformer-based models~\citep{schick-schutze-2021-exploiting,schick-schutze-2021-just,tam-etal-2021-improving,tunstall2022efficient,liu2022fewshot,karimi-mahabadi-etal-2022-prompt,aly-etal-2023-automated,albalak2023improving} or employing in-context learning (ICL) in Large Language Models (LLMs)~\citep{brown2020language,ahuja2023mega,asai2023buffet}. While few-shot benchmarks show clear progress in English, it is accompanied by both novel techniques and progress in the development of pre-trained models. For practitioners working with non-English data that advancement is always inspiring but does not fully answer the question: \emph{What approach or model should I use for a given task in language X?} This is grounded by the fact that different languages possess different resources (e.g. pre-trained models) and datasets. 

\begin{figure}[t]
    \centering
    \includegraphics[width=\linewidth]{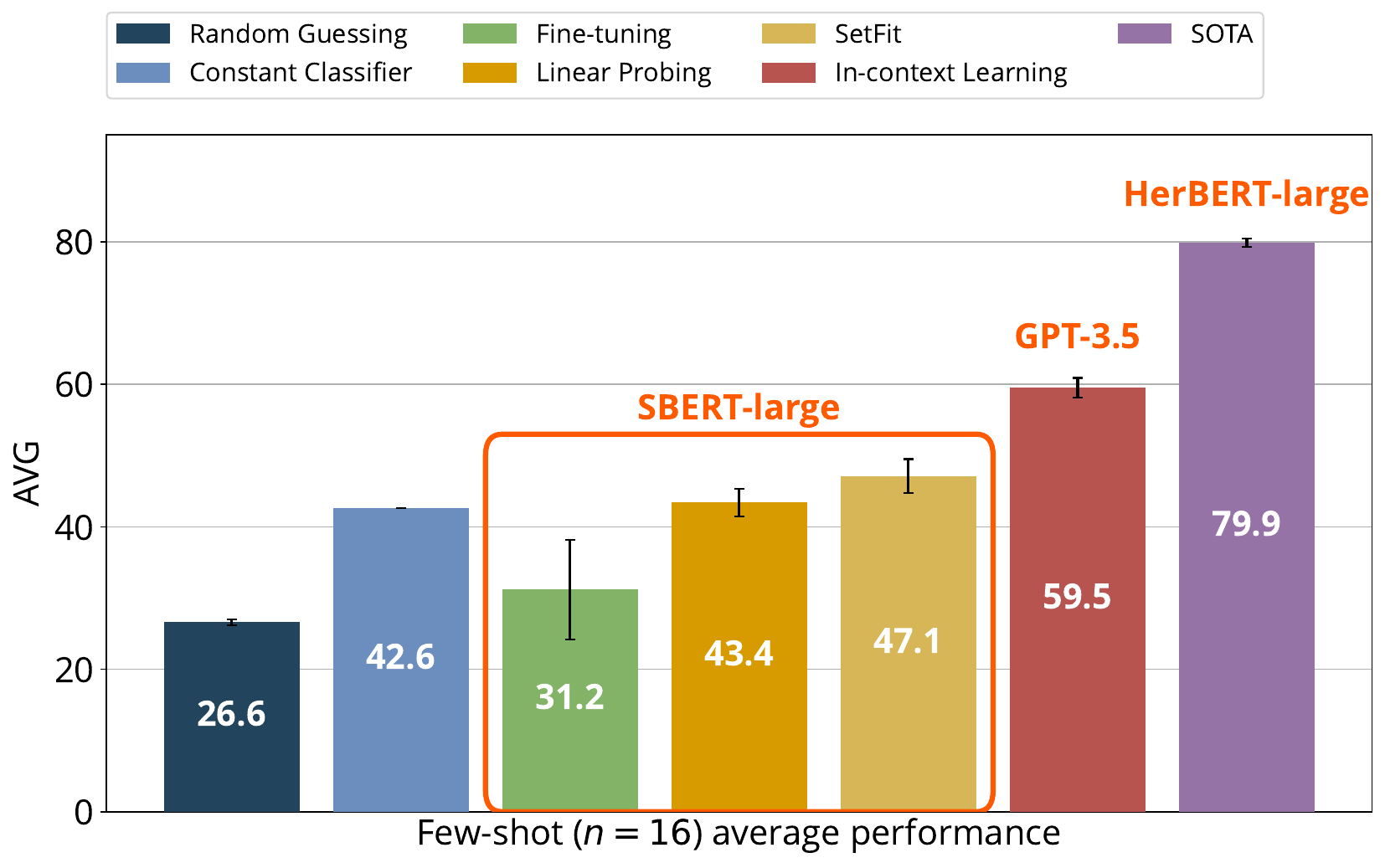}
    \caption{\textbf{Summary of average performance on the few-shot classification benchmark for different training techniques and 16 shots}. The last bar on the right reports results for the HerBERT-large fine-tuned on the whole train dataset. We observe that ICL with GPT-3.5 achieves the best average performance followed by SetFit, Linear probing, and fine-tuning with a much smaller SBERT-large model. The gap between HerBERT-large and GPT-3.5 is around 20.4 percentage points}
    \label{fig:summary}
\end{figure}

In this paper, we aim to evaluate various few-shot learning methodologies and assess the capability of pre-trained models in processing and understanding the Polish language. Unlike many studies that depend on multilingual or cross-lingual analysis with translated datasets, we emphasize native Polish language datasets to avoid potential translation and cultural biases~\citep{ruder-sil-2021-multi,artetxe-etal-2020-translation, ponti-etal-2020-xcopa}. We have developed a Polish language classification benchmark, incorporating 7 existing datasets: PAC, Polemo2, DYK, CDSC-E, NKJP-NER, CBD, and CST-Wikinews~\citep{NEURIPS2022_890b206e, rybak-etal-2020-klej}. These datasets were carefully chosen to represent a broad spectrum of characteristics, such as input length, domain specificity, and the number of classes. In our study, we used a few-shot benchmark to compare fine-tuning, linear probing, SetFit~\cite{tunstall2022efficient}, and in-context learning using various pre-trained models. Figure~\ref{fig:summary} summarizes the average performance for 16-shots. We hope this paper will benefit researchers focused on the Polish language and practitioners selecting suitable models for their tasks.

We summarize the contributions of this work as follows:
\begin{itemize}
    \item We utilized existing datasets to establish an evaluation benchmark comprising 7 different datasets for few-shot classification in the Polish language. We also created a collection of 71 manually crafted template prompts for in-context learning.
    \item We conducted a comparative study of fine-tuning, linear probing, SetFit, and ICL across various transformer-based models, revealing that ICL excels in highly scarce few-shot scenarios with only 16-shots. However, we observe a significant difference of 14.0 percentage points between the best-performing model, namely the zero-shot GPT-4, and the performance achieved by fine-tuning HerBERT-large on the entire dataset.
    \item We illustrate, with Bielik-7b and Trurl-2-13b, that continual pre-training in the target language is the zero-shot regime. For 16-shots, only Trurl-2-13b is better than its English counterpart. We also observe that continual pre-training on the target domain needs to be done with caution which is demonstrated by the lower performance of Krakowiak-7b-v2 which descends from Mistral-7b.
\end{itemize}

\begin{figure*}[!h]
	\centering
	\begin{subfigure}[b]{0.49\linewidth}
		\includegraphics[width=\textwidth]{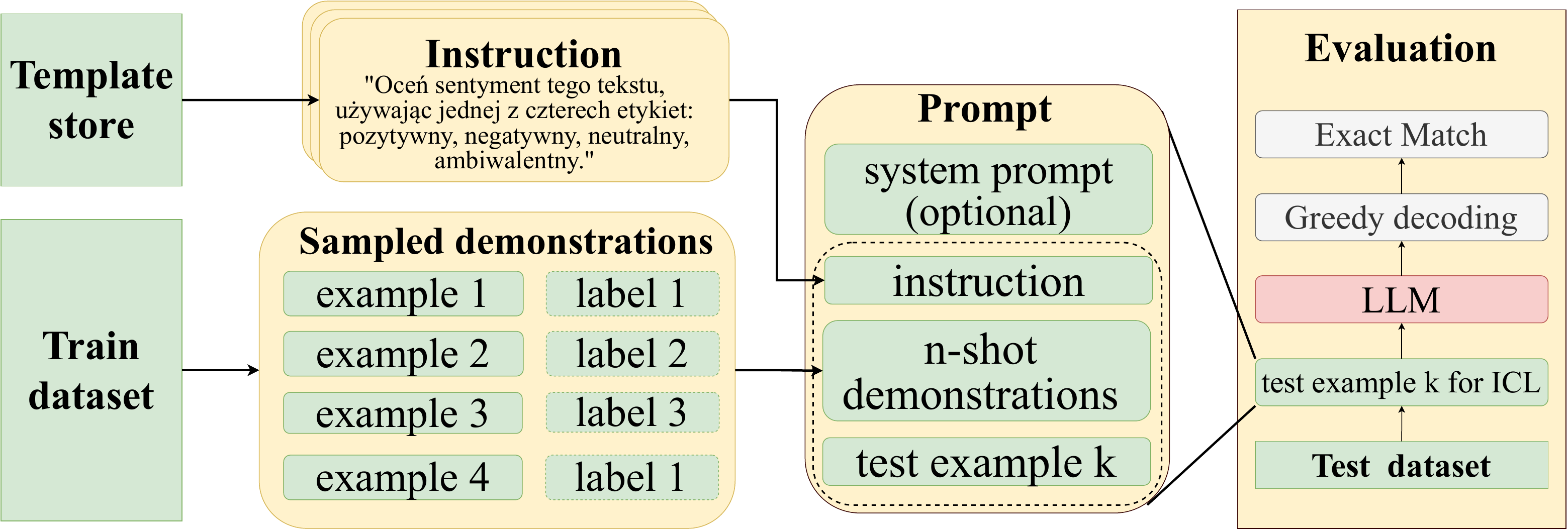}
		\caption{In-context learning evaluation}
	\end{subfigure}%
	\hfill
	\begin{subfigure}[b]{0.49\linewidth}
		\includegraphics[width=\textwidth]{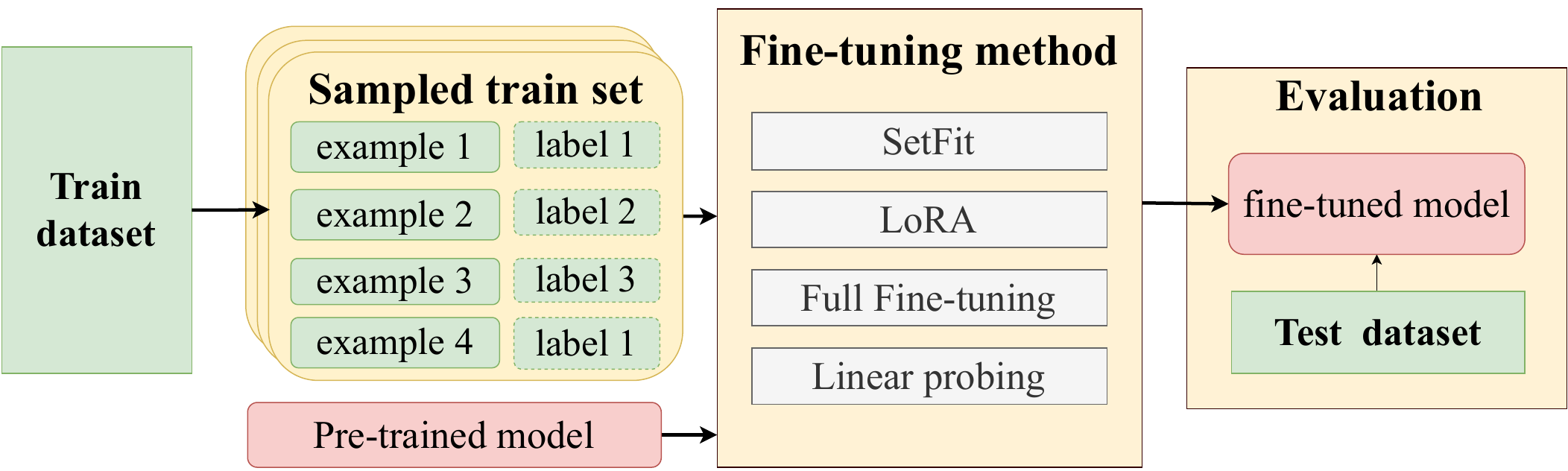}
		\caption{Fine-tuning methods evaluation}
	\end{subfigure}
	\caption{\textbf{Training and evaluation process of few-shot (as example 4-shot) learning}.
Both methods begin by sampling \(n\) (as \(n\)-shot) examples from the task's training data using a fixed random seed. We use 5 random seeds for reproducibility and to measure variance. 
\textbf{(a)} The sampled examples serve as in-context demonstrations. The evaluation prompt combines manually written instructions, demonstrations, and an optional system prompt. The model uses greedy decoding to generate a label for the test example. An exact match compares the generated label with the golden labels. Predictions without a matching label receive a special label.
\textbf{(b)} The sampled examples form the training dataset for the specified fine-tuning method (the Full Fine-tuning method utilizes the entire dataset). Subsequently, the fine-tuned model is evaluated on the test dataset in the conventional manner.
 }
	\label{fig:subfigures}
\end{figure*}

\section{Related Work}
\paragraph{Few-shot benchmarks}
Comparing models in a few-shot regime is not free from many nuances~\citep{bragg2021flex,ye-etal-2021-crossfit,alex2021raft,zheng-etal-2022-fewnlu}. The FewGLUE benchmark~\cite{schick-schutze-2021-just} randomly samples the same number of examples per task from the GLUE benchmark and provides several tens of thousands of unlabeled samples. Similarly, the CrossFit~\cite{ye-etal-2021-crossfit} uses a fixed number of examples per class. A small development set is also provided. The FLEX~\cite{bragg2021flex} argues that the benchmark should include a variable number of examples and lack of development set among other factors. The RAFT benchmark~\cite{alex2022raft} focuses on naturally occurring tasks and uses an evaluation setup that mirrors deployment.

\paragraph{Few-shot learning}
Over the years, numerous few-shot learning methods have been suggested. This discussion will focus on the most critical ones. First, ICL, used with large language models like GPT-3~\cite{brown2020language}, requires no modifications to the model. Demonstrations are joined with task instruction and forward through the model which generated the answer. As these models improve, they also become more effective at solving few-shot tasks. Another approach involves various forms of fine-tuning of pre-trained models. For instance, the PET~\cite{schick-schutze-2021-exploiting} and iPET~\citep{schick-schutze-2021-just} methods transform training examples into cloze questions and fine-tune the model on a small collection of examples, which is then used to softly annotate a pool of unlabeled examples. These methods rely on handcrafted task instructions and verbalizers, which the PERFECT method~\citep{karimi-mahabadi-etal-2022-prompt} replaces with task-specific adapters and does not depend on unlabeled data, similar to ADAPET~\citep{tam-etal-2021-improving}. Additionally, parameter-efficient fine-tuning has shown cutting-edge results as demonstrated in studies like~\citep{logan-iv-etal-2022-cutting}, T-Few~\citep{liu2022fewshot}, and AuT-Few~\citep{aly-etal-2023-automated}. Lastly, combining an embedding model with a linear classifier can yield good results with far fewer parameters, as seen with the SetFit method~\cite{tunstall2022efficient}, which begins with contrastive learning to enhance data comprehension and then trains a linear classifier on these refined, static embeddings.

\paragraph{Large Language Models (LLMs)}
In recent years, there has been significant growth in both commercial and open-source models~\citep{jiang2023mistral, touvron2023llama, gemmateam2024gemma}. Although many architectures are available~\citep{tay2023ul2}, the success of GPT-like models~\citep{brown2020language, ouyang2022training, openai2024gpt4} demonstrates the dominance of decoder-only architectures in generative tasks and ICL. Another key feature of modern models is their scale. Currently, there appears to be no limit to the benefits of scaling, leading to increasingly larger models~\citep{fedus2022switch, chowdhery2022palm, rae2022scaling, thoppilan2022lamda} and alternative scaling strategies such as Mixture-of-Experts (MoE)~\citep{shazeer2017outrageously,fedus2022switch, jiang2024mixtral}. Some research suggests that scaling endows models with new capabilities not present in smaller versions~\citep{wei2022emergent}, though this remains a topic of ongoing research~\citep{schaeffer2023emergent}. Another significant research area for LLMs involves fine-tuning methods~\citep{sanh2022multitask, chung2022scaling, wei2022finetuned} to enable few-shot learning and the use of natural language instructions. Reinforcement Learning (RL) methods~\citep{christiano2023deep, ouyang2022training} allows the integration of human feedback to improve the model's ability to generate coherent responses, although these methods are still challenging to train. New methods that could enhance or replace RL are being proposed~\citep{an2023direct, rafailov2023direct}.

\paragraph{Multilingual capabilities of LLMs}
Dominant studies on the performance of LLMs are done for the English language. Little was known about the multi-lingual capabilities of these models. Lately, a few papers studied this topic with both open-source and commercial models. ChatGPT's performance is generally better for tasks formulated in English than other languages~\citep{lai2023chatgpt,ahuja2023mega,zhang-etal-2023-dont,asai2023buffet}. However, a closer look at the data reveals that it strongly depends on the language and task. Typically, the performance gap is larger for low-resource languages. Following the same works, similar observations were made for the BLOOM~\cite{workshop2023bloom} and BLOOMZ~\cite{muennighoff-etal-2023-crosslingual} multi-lingual models. Analogous behavior is noticed for English-dominant models like Llama-2~\cite{bandarkar2023belebele}.

\section{Problem Statement}
Few-shot learning aims to learn a model from just a handful of labeled training examples. The special case is zero-shot learning where no labeled data is provided for training. We use $n$ to denote the number of total training examples.

Research in Polish NLP is lacking comparative evaluations among various few-shot techniques and pre-trained models. Additionally, the complexity of these methods and the computational resource requirements should be considered. Many non-English languages demonstrate cross-lingual capabilities but are often evaluated using translated datasets, which may introduce translation and cultural biases. Moreover, these methods are not routinely compared with simple baselines. In our study, we perform a comparative analysis of selected few-shot learning methods and pre-trained models on Polish classification tasks. We compare our results with both simple baselines and state-of-the-art (SOTA) results on the entire dataset. Our focus is on classification tasks due to their wide application across numerous fields and the simplicity of evaluation.

\begin{table*}[t]
\small
    \centering
    \begin{tabular*}{\linewidth}{C{3cm}|C{1.5cm}|C{1.5cm}|C{1.5cm}|C{1.7cm}|c|C{3cm}}
    \toprule
        \textbf{Name} & \textbf{Input} & \textbf{\#Test examples} & \textbf{\#Classes} & \textbf{\#Instructions} & \textbf{Avg. len} & \textbf{Dataset Domain} \\
        \midrule
        \multicolumn{7}{c}{Lepiszcze} \\
        \midrule
        PAC & text & 3453 & 2 & 13 & 185 & legal texts \\
        DYK & text pair & 1029 & 2 & 14 & 288 & Wikipedia \\
        CDSC-E & text pair & 1000 & 3 & 8 & 144 & image captions \\
        Polemo2 & text & 820 & 4 & 11 & 758 & online reviews \\
        \midrule
        \multicolumn{7}{c}{KLEJ} \\
        \midrule
        CBD & text & 1000 & 2 & 9 & 93 & social media \\
        NKJP-NER & text & 2058 & 6 & 9 & 85 & national corpus \\
        \midrule
        \multicolumn{7}{c}{Other} \\
        \midrule
        CST-Wikinews & text pair & 384 & 12 & 7 & 232 & Wikinews \\
        \bottomrule
    \end{tabular*}
    \caption{\textbf{Descriptions and basic statistics of datasets used as a benchmark for evaluation of few-shot learning capabilities}. The \#Instructions denote the number of handcrafted instructions for the in-context learning method. While creating a benchmark, we choose a dataset based on the parameters listed in the column. This helped us to prepare a diversified classification benchmark, with data from different domains, different number of classes, and length distribution.}
    \label{tab:datasets}
\end{table*}

\section{Methodology}
\subsection{Datasets}
Numerous classification datasets are available for the Polish language. For the construction of the benchmark, 7 such datasets were selected, which vary in domain and class quantity. The PAC, Polemo2, DYK, and CDSC-E datasets constitute a subset of the Lepiszcze benchmark~\cite{NEURIPS2022_890b206e}. Meanwhile, the NKJP-NER and CBD datasets are incorporated within the KLEJ benchmark~\cite{rybak-etal-2020-klej}. Additionally, the CST-Wikinews dataset comprises 1,530 training sentence pairs extracted from Polish Wikipedia, annotated with 12 distinct relational categories between the first and second sentences, such as inclusion, description, and summarization~\footnote{\url{https://huggingface.co/datasets/clarin-pl/cst-wikinews}}. The datasets also exhibit variation in the average length of examples. Table \ref{tab:datasets} provides a summary of these datasets. Links to open-source artifacts can be found in the Appendix~\ref{app:datasets}.

\paragraph{Evaluation metrics} For datasets endowed with binary labels, the F1 score for the positive class is reported, whereas Accuracy is employed for the other datasets. Such a combination of metrics is used to have consistency with the results from the KLEJ benchmark~\cite{rybak-etal-2020-klej}.

\begin{table*}[!h]
\addtolength\tabcolsep{-2.5pt}
\renewcommand{\arraystretch}{1.2}
\small
    \centering
    \begin{tabular}{c|c|cccccccc}
    \toprule
        & \rot{AVG} & \rot{PAC} & \rot{Polemo2} & \rot{CBD} & \rot{NKJP-NER} & \rot{DYK} & \rot{CDSC-E} & \rot{CST-Wikinews} \\
        \midrule
        XGBoost & 32.4\down{ ± 3.1} & 56.4\down{ ± 16.7} & 38.5\down{ ± 3.4} & 19.7\down{ ± 6.0} & 20.9\down{ ± 4.7} & 22.6\down{ ± 5.1} & 57.5\down{ ± 10.4} & 11.4\down{ ± 1.5} \\
        Random Guessing & 26.6\down{ ± 0.4} & 57.5\down{ ± 0.7} & 24.8\down{ ± 1.0} & 20.9\down{ ± 1.6} & 16.8\down{ ± 0.8} & 25.5\down{ ± 1.7} & 33.6\down{ ± 0.7} & 6.9\down{ ± 0.9} \\
        Constant Classifier & 42.9 & 80.6 & 41.3 & 23.6 & 34.3 & 28.9 & 74.4 & 16.9 \\
        \midrule
        /SF/ SBERT-large & 47.1\down{ ± 2.4} & 68.8\down{ ± 6.5} & 69.9\down{ ± 10.7} & 44.4\down{ ± 5.4} & 30.7\down{ ± 6.5} & 27.7\down{ ± 2.9} & 72.3\down{ ± 5.7} & 16.2\down{ ± 3.2} \\
        /LP/ SBERT-large & 43.4\down{ ± 1.9} & 67.5\down{ ± 7.0} & 60.3\down{ ± 4.6} & 40.1\down{ ± 4.3} & 30.4\down{ ± 5.9} & 27.2\down{ ± 1.8} & 62.8\down{ ± 5.8} & 15.3\down{ ± 3.0} \\
        /FT/ SBERT-large & 31.2\down{ ± 7.0} & 33.6\down{ ± 31.3} & 47.0\down{ ± 5.8} & 32.0\down{ ± 30.7} & 28.1\down{ ± 7.7} & 6.4\down{ ± 10.3} & 61.0\down{ ± 17.6} & 10.1\down{ ± 4.3} \\
        \midrule
        /ICL\down{n=16}/ GPT-3.5 & 59.5\down{ ± 1.4} & 73.9\down{ ± 3.6} & 81.9\down{ ± 2.1} & \textbf{64.1}\down{ ± 1.9} & 46.1\down{ ± 2.9} & 64.1\down{ ± 1.8} & 66.7\down{ ± 7.6} & \textbf{19.8}\down{ ± 2.7} \\
        /ICL\down{n=0}/ GPT-4 & \textbf{65.9} & 83.3 & \textbf{82.6} & 60.6 & \textbf{58.8} & \textbf{81.3} & \textbf{76.0} & 18.8 \\
        \midrule
        HerBERT-large & 79.9\down{ ± 0.6} & 91.1\down{ ± 0.0} & 90.9\down{ ± 0.0} & 53.2\down{ ± 3.2} \textcolor{red}{(72.0)} & 94.0\down{ ± 0.0} & 68.8\down{ ± 2.1} \textcolor{red}{(75.8)} & 93.4\down{ ± 0.0} & 67.9\down{ ± 1.0} \\
        \bottomrule
    \end{tabular}
    \caption{\textbf{Evaluation results summary on the test dataset}. We employed a dataset comprising 16 training examples per task, utilizing 5 distinct random seeds for analysis. Subscripts denote standard deviation. The acronyms SF, LP, FT, and ICL represent SetFit, Linear Probing, Fine-Tuning, and In-context Learning, respectively. The outcomes for the GPT model family pertain to in-context learning. For GPT-4 model we report zero-shot performance. The terminal row delineates the performance metrics after fine-tuning across the entire training dataset. Parenthetically, we reference the results derived from the KLEJ benchmark~\footnote{\url{https://klejbenchmark.com/leaderboard/}}.}
    \label{tab:summary}
\end{table*}

\subsection{Experimental setup}
We adopt a methodology analogous to that described in~\cite{schick-schutze-2021-just}. 
For each dataset, we sample a random subset comprising $n=16$ instances from the training set. We wanted the number to be as maximal as possible and still possible to process by the ICL method. It is critical to ensure that the derived subset is balanced (where feasible) and that the labels are sequenced in an alternating fashion. The entirety of the validation dataset is utilized for the hyperparameter selection. In the absence of a predefined validation set, the training set is partitioned into a 20\% validation subset and an 80\% training subset. Consequently, our analysis concentrates on the upper-performance limit in few-shot scenarios, given the disproportionately large size of the validation dataset. The experiment is conducted 5 times with varying random subsets. We set the following random seeds in our experiments: [18, 22, 37, 69, 98]. For each approach, the arithmetic mean and standard deviation are documented. Additionally, for the ICL, we also report the zero-shot performance.
\subsection{Models}
Our study concentrates on the evaluation of transformer-based models with specific attributes, encompassing: (a) medium-sized models, such as Polish Sentence BERT-large and HerBERT-large; (b) Polish open-source LLMs, notably GPT-2-xl, Trurl-2, Bielik-7b, and Krakowiak-v2; (c) English open-source LLMs, such as Llama-2 and Mistral-7b; and (d) commercial LLMs, including PaLM-2, GPT-3.5, and GPT-4. The selection criteria are predicated on language support and the incorporation of instruction tuning~\citep{mishra-etal-2022-cross,sanh2022multitask,wei2022finetuned} during the pretraining phase. Details on model artifacts can be found in the Appendix~\ref{app:models}.

\subsection{Training schemes}
The experiments conducted can be divided into two categories: few-shot ICL and few-shot parameter-efficient learning. The former focuses on generating labels using only demonstrations and task descriptions in the input, without updating the model weights. The latter involves fine-tuning the model for a new task using only $n$ examples. These approaches are depicted in Figure~\ref{fig:subfigures}. Below, we provide more detailed descriptions for each method.

\paragraph{Baseline}
The baseline model comprises an \emph{XGBoost}~\cite{Chen:2016:XST:2939672.2939785} algorithm, trained utilizing hashed Term Frequency-Inverse Document Frequency (TF-IDF) features. Initially, text tokenization is conducted via \texttt{spaCy}~\cite{spacy2}. Subsequently, each token is hashed employing the Fowler-Noll-Vo (FNV) algorithm and then mapped to a predetermined vocabulary dimension. Performance metrics are also presented for \emph{Random Guessing}, entailing random label selection, and for the \emph{Constant Classifier} method, which consistently predicts the label maximizing the test score.

\paragraph{Linear probing}
In this approach, each example is projected onto a constant embedding using a pre-trained model, followed by training a Logistic Regression classifier on these embeddings. During the training process, embeddings are frozen and not normalized.

\paragraph{SetFit}
SetFit represents an effective fine-tuning methodology for few-shot learning~\citep{tunstall2022efficient}, combining metric learning with linear probing. The implementation of SetFit is conducted in conjunction with LoRA~\cite{hu2022lora} applied to all linear layers.

\paragraph{Fine-tuning}
The fine-tuning of the backbone transformer model is executed using cross-entropy loss, integrating a non-linear classification head. Throughout the training phase, only the LoRA modules (applied to all linear layers) alongside the head parameters are adjusted. The same learning rate is applied for both the backbone and the head components. 

\paragraph{In-context learning}
In-context learning functions without the alteration of model parameters~\citep{radford2019language,brown2020language}. Instructions and examples are provided as input to the language generation model, from which an appropriate label is anticipated as generated output. Further details are available in Appendix~\ref{app:in-context-learning}.

\subsection{Prompt Engineering}
For each dataset, we systematically hand-crafted diverse prompts derived from the dataset characteristics and anticipated label categories. We mostly followed PromptSource~\cite{bach-etal-2022-promptsource} and Natural Instructions~\cite{mishra-etal-2022-cross,wang-etal-2022-super}, without employing automated prompt generation methods~\citep{zhou2023large,deng-etal-2022-rlprompt,shin-etal-2020-autoprompt,prasad-etal-2023-grips,pryzant-etal-2023-automatic,anonymous2024connecting,anonymous2024large,fernando2023promptbreeder}. The construction of each prompt entailed 3 distinct segments. It starts with universal system directive in Polish, specifically: \textit{"Rozwiązujesz zadanie klasyfikacji dla języka polskiego."}. This was followed by optional task-specific instruction and subsequent demonstrations. Refer to Figure~\ref{fig:prompt-template} for an exhaustive depiction. The prompts are cataloged in the Appendix~\ref{app:templates}.

\begin{figure}[!h]
\small\begin{minted}[
    frame=single,
    linenos,
    breaklines
]{yaml}
instruction: 'Oceń czy następujący zapis jest uczciwy lub nieuczciwy.'
demonstration: '{text} => {labels}'
label_mapping:
    0: nieuczciwy
    1: uczciwy
\end{minted}
\caption{\textbf{Example building blocks of the prompt for in-context learning}. Every prompt start with system instruction: \textit{"Rozwiązujesz zadanie klasyfikacji dla języka polskiego."}. Then, follows a \textbf{instruction} and a sequence of demonstrations. Every block is joined with a newline character. The \textbf{demonstration} part specifies the way to format examples. Specifically, \texttt{\{text\}} and \texttt{\{labels\}} are substituted with text and true labels from the dataset, respectively. Label mapping is used to map textual labels into numeric values.}
\label{fig:prompt-template}
\end{figure}

\paragraph{Response cleansing}
In the domain of ICL, each model produces a response that is subsequently classified into one of the predefined labels using some heuristics. We assume that the label is present in the response.
Minimal postprocessing is employed, encompassing conversion to lowercase, removal of diacritics, and elimination of extraneous whitespace. There was no pursuit of more advanced methodologies, such as instructing language models to generate outputs conforming to a specific JSON schema.
If the label is absent in the response, it is treated as invalid. When we calculate the metrics, we replace invalid responses with a class different from ground truth to lower the metrics automatically.

\section{Results \& Discussion}

Table~\ref{tab:summary} summarizes the outcomes from various training methodologies. The HerBERT-large model, fine-tuned on the entire training dataset using Focal Loss~\citep{8417976}\footnote{Additional experiments were conducted with scripts from the Lepiszcze benchmark \url{https://github.com/dkajtoch/LEPISZCZE}, resulting in similar outcomes.}, is also included. Scores from the KLEJ benchmark for the CBD and DYK tasks are shown in parentheses~\footnote{\url{https://klejbenchmark.com/leaderboard/}}.

We found that nearly all models outperform the simple XGBoost baseline significantly. An exception is the fine-tuning performance on PAC and DYK tasks, which lags behind. Additionally, almost all models surpass the Random Guessing baseline, including XGBoost. However, the Constant Classifier sets a notable threshold due to the severe imbalance in the PAC and CDSC-E datasets, which all models except GPT-4 fail to exceed. For the other datasets, the average performance is better than the Constant Classifier baseline.

ICL with commercial LLMs achieves the highest performance across the models and techniques. The second-best method is SetFit, followed by Linear probing. The poorest performance and greatest uncertainties are associated with fine-tuning. We note that linear probing is the most computationally efficient method. On the other hand, ICL requires large models and is computationally expensive. 

ICL using commercial models like GPT-4 shows superior performance. However, there is a gap of 14pp between zero-shot GPT-4 and HerBERT-large trained on the full dataset. Notably, GPT-4 surpasses comprehensive fine-tuning on the DYK task by 5.5 percentage points, reaching 81.3 points. This discrepancy warrants further investigation into potential dataset leakage, as GPT-3.5 shows much lower performance.

\subsection{Linear probing}
Table~\ref{tab:results_embeddings} presents the outcomes of employing a linear probing methodology across various embedding models. The investigation prioritized 5 open-source and 4 commercial backbones. SBERT-large exhibits exceptional efficacy in the CBD and NKJP-NER tasks, securing the top positions within the linear probing models for these datasets. For the DYK and CDSC-E datasets, HerBERT-large is identified as the most effective performer in this segment. 

Among the linear probing models, no single model consistently surpasses the rest across all tasks; however, SBERT-large and HerBERT-large demonstrate robust performance in distinct datasets. When considering the standard deviation, the performance of all models tends to converge on an average scale. Nonetheless, a pronounced disparity is observed in the Polemo2 task, where SBERT-large significantly outshines HerBERT-large, yet shows comparable performance to RoBERTa-large.

On average, the Gecko embeddings underperform relative to other backbones, exhibiting performance akin to that of the Llama-2-70b model. This finding starkly contrasts with the superior average performance achieved by the multilingual variant of the Gecko embeddings~\footnote{Determining whether the enhanced performance of the multilingual version is attributable to linguistic factors is complex. Both versions are predicated on the identical Gecko iteration but were released at different times. Subsequent versions also underwent API modifications and necessitated specifying \texttt{task\_name=CLASSIFICATION}.}.

Embedding-based few-shot learning is generally plagued by high variance, occasionally escalating to 12 percentage points. This variance stems from the sampling of demonstrations and the fine-tuning process. In the context of embeddings, the validation dataset offers limited utility.

\begin{table*}[!h]
\addtolength\tabcolsep{-2.5pt}
\renewcommand{\arraystretch}{1.2}
\small
    \centering
    \begin{tabular}{c|c|cccccccc}
    \toprule
        & \rot{AVG} & \rot{PAC} & \rot{Polemo2} & \rot{CBD} & \rot{NKJP-NER} & \rot{DYK} & \rot{CDSC-E} & \rot{CST-Wikinews} \\
        \midrule
        SBERT-large & 43.4\down{ ± 1.9} & 67.5\down{ ± 7.0} & 60.3\down{ ± 4.6} & \underline{40.1\down{ ± 4.3}} & \underline{30.4\down{ ± 5.9}} & 27.2\down{ ± 1.8} & 62.8\down{ ± 5.8} & 15.3\down{ ± 3.0} \\
        HerBERT-large & 43.0\down{ ± 2.4} & 72.6\down{ ± 10.3} & 47.0\down{ ± 3.1} & 33.8\down{ ± 10.1} & 27.8\down{ ± 3.3} & \underline{34.6\down{ ± 2.8}} & \underline{71.0\down{ ± 6.1}} & 14.2\down{ ± 2.7} \\
        RoBERTa-large & 41.4\down{ ± 2.4} & 67.8\down{ ± 11.3} & 57.9\down{ ± 4.4} & 37.4\down{ ± 6.1} & 27.6\down{ ± 3.9} & 25.1\down{ ± 3.0} & 59.2\down{ ± 7.5} & 14.5\down{ ± 2.7}  \\
        GPT-2-large & 39.8\down{ ± 1.7} & 72.0\down{ ± 2.9} & 48.3\down{ ± 4.6} & 36.4\down{ ± 4.1} & 25.2\down{ ± 4.3} & 28.6\down{ ± 3.1} & 54.4\down{ ± 7.5} & 13.6\down{ ± 3.0} \\
        Llama-2-70b\raise0.5ex\hbox{*} & 38.1\down{ ± 2.2} & 67.6\down{ ± 12.5} & 41.4\down{ ± 2.9} & 28.4\down{ ± 2.5} & 24.2\down{ ± 2.0} & 30.7\down{ ± 3.0} & 61.1\down{ ± 6.1} & 13.5\down{ ± 4.2}\\
        \midrule
        Ada & 40.9\down{ ± 2.3} & \underline{72.9\down{ ± 6.3}} & 55.2\down{ ± 5.0} & 30.7\down{ ± 4.2} & 29.1\down{ ± 2.5} & 25.4\down{ ± 4.1} & 58.2\down{ ± 11.7} & 14.8\down{ ± 3.9}\\
        DaVinci & 42.7\down{ ± 2.6} & 67.6\down{ ± 7.0} & 58.9\down{ ± 1.3} & 36.9\down{ ± 9.7} & 30.1\down{ ± 6.6} & 29.9\down{ ± 2.7} & 60.1\down{ ± 11.2} & 15.5\down{ ± 3.5}\\
        Gecko & 37.7\down{ ± 1.7} & 61.8\down{ ± 4.5} & 42.9\down{ ± 6.4} & 23.7\down{ ± 1.3} & 24.8\down{ ± 4.3} & 25.0\down{ ± 3.8} & 68.8\down{ ± 6.3} & \underline{17.0\down{ ± 2.8}}\\
        Gecko multilingual & 44.0\down{ ± 2.3} & 72.2\down{ ± 5.6} & \underline{60.8\down{ ± 2.6}} & 41.0\down{ ± 3.3} & 28.4\down{ ± 2.8} & 27.4\down{ ± 2.3} & 62.1\down{ ± 12.6} & 16.2\down{ ± 5.7}\\
        \bottomrule
    \end{tabular}
    \caption{\textbf{Evaluation results on the test dataset for linear probing}. We used 16 train examples per task and 5 random seeds. Subscripts denote standard deviation. \raise0.5ex\hbox{*}Llama-2 embeddings were generated with 4-bit quantization.}
    \label{tab:results_embeddings}
\end{table*}

\subsubsection{Scaling}
\begin{figure}[t]
    \centering
    \includegraphics[width=\linewidth]{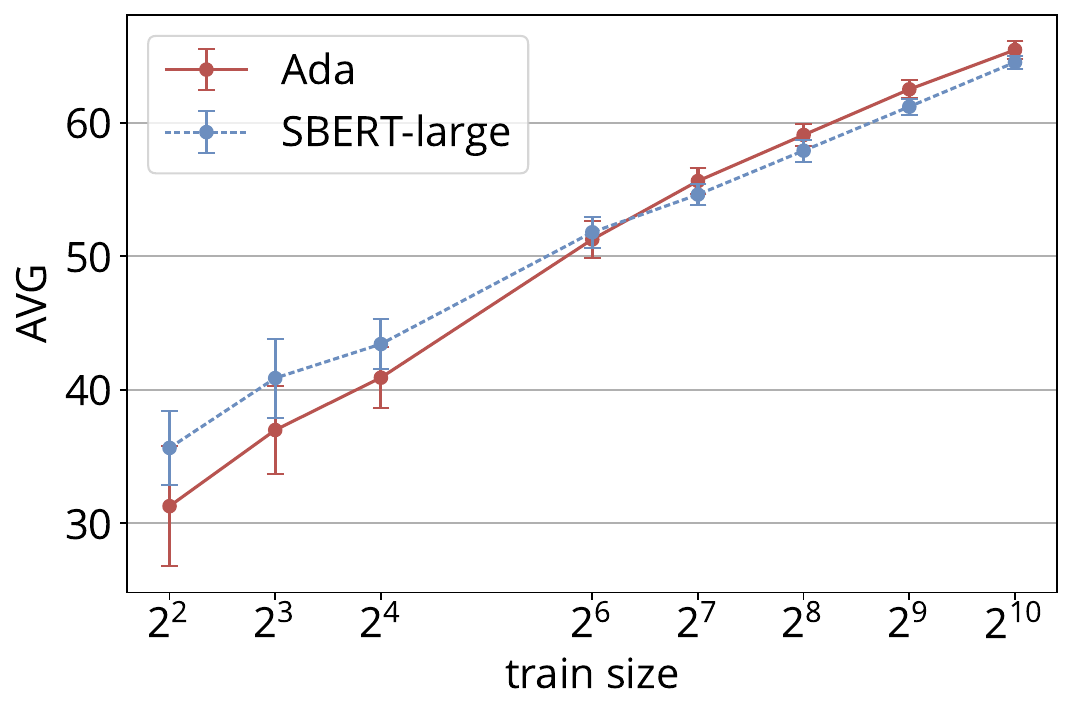}
    \caption{Average performance of the linear probing method based on Ada and SBERT-large embeddings as a function of train dataset size.}
    \label{fig:scaling_linear_probing}
\end{figure}

Linear probing represents a straightforward and cost-efficient method among the available techniques. This elicits the question, \textit{How many examples are required to achieve a predetermined performance level?} In Figure~\ref{fig:scaling_linear_probing}, the average performance of SBERT-large and Ada across various training example counts is illustrated. An almost perfect logarithmic relationship is observed between the average score and the quantity of examples. With the increment in the number of examples, the uncertainty diminishes, and Ada exhibits marginally superior performance compared to the SBERT-large model.

\subsection{Fine-tuning}
Table~\ref{tab:results_fine_tuning} presents the results of fine-tuning for $n=16$ training instances across various backbones. Generally, achieving a stable model using the fine-tuning approach proved to be challenging, leading to ambiguous comparisons between backbone efficacies. Conversely, SetFit demonstrated consistent performance across HerBERT-based backbones. In the context of this limited training scenario, the superiority of Sentence BERT over the original BERT remains uncertain due to significant variability. It is imperative to highlight that both fine-tuning methodologies employed LoRA approximation across all linear layers, necessitating cautious interpretation compared with comprehensive parameter fine-tuning.

\begin{table*}[!h]
\addtolength\tabcolsep{-2.5pt}
\renewcommand{\arraystretch}{1.2}
\small
    \centering
    \begin{tabular}{c|c|cccccccc}
    \toprule
        & \rot{AVG} & \rot{PAC} & \rot{Polemo2} & \rot{CBD} & \rot{NKJP-NER} & \rot{DYK} & \rot{CDSC-E} & \rot{CST-Wikinews} \\
        \midrule\midrule
        \multicolumn{9}{c}{with non-linear head} \\
        \midrule
        SBERT-large & 31.2\down{ ± 7.0} & 33.6\down{ ± 31.3} & 47.0\down{ ± 5.8} & 32.0\down{ ± 30.7} & 28.1\down{ ± 7.7} & 6.4\down{ ± 10.3} & 61.0\down{ ± 17.6} & 10.1\down{ ± 4.3} \\
        HerBERT-large & 25.9\down{ ± 8.6} & 26.7\down{ ± 23.4} & 29.5\down{ ± 9.7} & 30.8\down{ ± 36.6} & 25.2\down{ ± 3.8} & 16.2\down{ ± 29.7} & 44.4\down{ ± 26.7} & 8.7\down{ ± 3.7} \\
        RoBERTa-large & 24.2\down{ ± 4.5} & 6.2\down{ ± 11.5} & 36.4\down{ ± 4.3} & 15.3\down{ ± 10.5} & 21.3\down{ ± 6.1} & 27.2\down{ ± 0.4} & 54.5\down{ ± 26.0} & 8.7\down{ ± 2.8} \\
        \midrule\midrule
        \multicolumn{9}{c}{SetFit} \\
        \midrule
        SBERT-large & 47.1\down{ ± 2.4} & 68.8\down{ ± 6.5} & 69.9\down{ ± 10.7} & 44.4\down{ ± 5.4} & 30.7\down{ ± 6.5} & 27.7\down{ ± 2.9} & 72.3\down{ ± 5.7} & 16.2\down{ ± 3.2} \\
        HerBERT-large & 44.1\down{ ± 3.9} & 70.7\down{ ± 11.9} & 46.0\down{ ± 11.2} & 42.5\down{ ± 10.9} & 25.8\down{ ± 11.6} & 40.6\down{ ± 4.7} & 67.2\down{ ± 14.5} & 15.7\down{ ± 1.9} \\
        \bottomrule
    \end{tabular}
    \caption{\textbf{Evaluation results for the fine-tuning technique on the test dataset}. We used 16 train examples per task and 5 random seeds. Subscripts denote standard deviation. All models used LoRA approximation across all linear layers.}
    \label{tab:results_fine_tuning}
\end{table*}

\subsection{In-context learning}
Prior methodologies concentrated on models characterized by a constrained set of hyperparameters, such as learning rate, batch size, and training steps. In contrast, ICL employs pre-trained models that necessitate no further training, yet a text prompt furnishes an exponential array of choices. Presently, it is feasible to differentiate between models based on various attributes, including size, corpus language, and data formats employed for fine-tuning, such as instruction tuning.

\paragraph{Few-shot}
Table~\ref{tab:icl_few_shot} provides a detailed comparative analysis of the few-shot ($n=16$) ICL performance of various language models across distinct classification tasks. Commercially developed models such as GPT-3.5, achieving an average score of 59.5, and Bison-text, with 53.4, exhibit good performance metrics. GPT-3.5, in particular, demonstrates superior capability, notably in the Polemo2 task, achieving an 81.9 score. Conversely, Bison-text does very well on the PAC task, scoring 83.7. However, it does poorly on the CBD task, getting a 0.0 score. This poor performance is because it fails to produce suitable responses. This failure is linked to breaking Google's rules about content related to cyberbullying, which includes inappropriate language.

Further exploration into task-specific performances reveals the Polemo2 task as a significant challenge, largely due to the constraints related to the models' context size. This constraint leads to zero scores in models such as Bielik-7b-instruct, Trurl-2-13b, and GPT-2-xl, each posting an average performance score of 31.7, 22.5, and 19.3, respectively. The considerable uncertainty metrics, like ±12.0 for Trurl’s PAC task and ±22.5 for Krakowiak's PAC, underscore the variability and challenges inherent in securing stable few-shot learning outcomes. In stark contrast, Mistral-7b-instruct, though fine-tuned for English, attains a moderate average score of 46.5, with a notably high score of 77.6 in the CDSC-E task, indicating potential cross-lingual learning capabilities. Llama-2-13b-chat, on the other hand, significantly underperforms with an average score of merely 12.2. Mistral-7b-instruct emerges as the superior model among open-source variants, despite not being specifically fine-tuned for Polish, reflecting the complex dynamics of few-shot learning in a multilingual environment where language-specific fine-tuning and intrinsic model capabilities are crucial. Bielik-7b-instruct which is based on the same architecture and initial weights attains significantly worse performance. Further analysis is needed to understand Mistral's performance.  

\begin{table*}[!h]
\sethlcolor{lightyellow}
\addtolength\tabcolsep{-2.5pt}
\renewcommand{\arraystretch}{1.2}
\small
    \centering
    \begin{tabular}{c|c|ccccccc}
    \toprule
        & \rot{AVG} & \rot{PAC} & \rot{Polemo2} & \rot{CBD} & \rot{NKJP-NER} & \rot{DYK} & \rot{CDSC-E} & \rot{CST-Wikinews}\\
        \midrule
        GPT-3.5 & 59.5\down{ ± 1.4} & 73.9\down{ ± 3.6} & 81.9\down{ ± 2.1} & 64.1\down{ ± 1.9} & 46.1\down{ ± 2.9} & 64.1\down{ ± 1.8} & 66.7\down{ ± 7.6} & 19.8\down{ ± 2.7} \\
        Bison-text & 53.4\down{ ± 1.3} & 83.7\down{ ± 1.4} & 81.8\down{ ± 2.8} & 0.0\down{ ± 0.0} & 45.7\down{ ± 3.6} & 76.6\down{ ± 0.7} & 66.4\down{ ± 7.3} & 19.5\down{ ± 2.2} \\
        \midrule
        Bielik-7b-instruct & 31.7\down{ ± 1.5} & 76.1\down{ ± 4.7} & \hl{0.0}\down{ ± 0.0} & 23.7\down{ ± 0.1} & 23.2\down{ ± 2.2} & 38.3\down{ ± 3.8} & 55.0\down{ ± 10.7} & 5.4\down{ ± 1.3} \\
        Trurl-2-13b & 22.5\down{ ± 4.0} & 66.7\down{ ± 12.0} & \hl{0.0}\down{ ± 0.0} & 12.1\down{ ± 6.6} & 15.9\down{ ± 11.3} & 30.8\down{ ± 1.4} & 26.1\down{ ± 21.8} & 6.3\down{ ± 3.2} \\
        Krakowiak-7b-v2 & 22.5\down{ ± 5.5} & 57.9\down{ ± 22.5} & 14.4\down{ ± 18.5} & 0.0\down{ ± 0.0} & 13.8\down{ ± 1.0} & 28.9\down{ ± 0.1} & 38.0\down{ ± 24.5} & 4.8\down{ ± 4.7} \\
        GPT-2-xl & 19.3\down{ ± 4.0} & 24.8\down{ ± 24.7} & \hl{0.0}\down{ ± 0.0} & 0.0\down{ ± 0.0} & 14.0\down{ ± 11.2} & 16.9\down{ ± 1.8} & 74.0\down{ ± 0.5} & 5.2\down{ ± 5.7} \\
        \midrule
        Mistral-7b-instruct & 46.5\down{ ± 2.1} & 48.8\down{ ± 6.8} & 73.5\down{ ± 6.1} & 34.4\down{ ± 8.6} & 28.2\down{ ± 1.8} & 45.1\down{ ± 1.6} & 77.6\down{ ± 6.5} & 17.9\down{ ± 2.4} \\
        Llama-2-13b-chat & 12.2\down{ ± 6.8} & 47.8\down{ ± 41.7} & \hl{0.0}\down{ ± 0.0} & 0.2\down{ ± 0.4} & 3.9\down{ ± 7.7} & 20.0\down{ ± 8.8} & 12.1\down{ ± 18.9} & 1.1\down{ ± 2.4}\\
        \bottomrule
    \end{tabular}
    \caption{\textbf{Evaluation results for the ICL on the test dataset}. We used 16 train examples per task and 5 random seeds. Subscripts denote standard deviation. Highlighted scores for the Polemo2 task are 0.0 since 16 demonstrations do not fit into the context of selected models.}
    \label{tab:icl_few_shot}
\end{table*}

\begin{table*}[!h]
\addtolength\tabcolsep{-2.5pt}
\renewcommand{\arraystretch}{1.2}
\small
    \centering
    \begin{tabular}{c|c|ccccccc}
    \toprule
        & \rot{AVG} & \rot{PAC} & \rot{Polemo2} & \rot{CBD} & \rot{NKJP-NER} & \rot{DYK} & \rot{CDSC-E} & \rot{CST-Wikinews}\\
        \midrule
        GPT-4 & \textbf{65.9} & \textbf{83.3}\down{.0} & \textbf{82.6}\down{.009} & \textbf{60.6}\down{.028} & \textbf{58.8}\down{.004} & \textbf{81.3}\down{.002} & \textbf{76.0}\down{.0} & \textbf{18.8}\down{.003} \\
        GPT-3.5 & 55.4 & 82.2\down{.014} & 81.6\down{.005} & 50.0\down{.046} & 44.9\down{.001} & 53.1\down{.0} & 62.9\down{.0} & 13.3\down{.003} \\
        Bison-text & 52.2 & 80.2\down{.006} & 80.7\down{.009} & 42.6\down{.077} & 47.5\down{.027} & 61.6\down{.016} & 35.0\down{.001} & 17.7\down{.003} \\
        \midrule
        Bielik-7b-instruct & 51.5 & 78.1\down{.034} & 77.3\down{.0} & 53.9\down{.0} & 31.0\down{.01} & 49.0\down{.001} & 65.3\down{.0} & 6.0\down{.023} \\
        Trurl-2-13b & 50.2 & 74.1\down{.064} & 79.5\down{.001} & 50.0\down{.001} & 33.6\down{.038} & 49.7\down{.015} & 50.2\down{.001} & 14.6\down{.083}\\
        Krakowiak-7b-v2 & 43.9 & \hl{80.6}\down{.001} & 65.0\down{.018} & \hl{23.6}\down{.0} & 16.7\down{.024} & 41.4\down{.002} & \hl{74.4}\down{.0} & 6.0\down{.034} \\ 
        GPT-2-xl & 13.7 & 27.0\down{.619} & 5.1\down{.827} & 2.9\down{.794} & 13.3\down{.513} & 18.9\down{.3} & 29.1\down{.519} & 0.0\down{.992} \\

        \midrule
        Mistral-7b-instruct & 50.7 & 78.8\down{.036} & 70.0\down{.043} & 40.2\down{.001} & 36.8\down{.054} & 52.7\down{.001} & 65.2\down{.0} & 11.7\down{.078} \\
        Llama-2-13b-chat & 33.7 & 56.1\down{.069} & 62.7\down{.048} & 10.1\down{.508} & 4.6\down{.785} & 37.3\down{.024} & 63.1\down{.029} & 1.8\down{.797} \\
        \bottomrule
    \end{tabular}
    \caption{\textbf{Zero-shot evaluation results for the ICL on the test dataset}. The subscript indicates a fraction of the test set with invalid predictions. We only provide results for templates with a minimal number of invalid predictions. Highlighted scores (orange) denote the same performance as the Constant Classifier.}
    \label{tab:zero-shot}
\end{table*}

\paragraph{Zero-shot}
Table~\ref{tab:zero-shot} presents a comparative analysis of zero-shot in-context learning performance across various language models on different classification tasks, revealing significant disparities. GPT-4 emerges as the clear leader, boasting the highest average performance of 65.9 points and leading in every individual task. Following GPT-4, GPT-3.5 shows solid performance with an overall average of 55.4. Models like Bison-text, Bielik-7b-instruct, and Mistral-7b-instruct demonstrate competitive capabilities, with averages around 52.2, 51.5, and 50.7 respectively, while Krakowiak-7b-v2 shows variable performance, excelling in specific tasks but having a lower overall average. Despite Krakowiak's commendable scores, its performance on tasks such as PAC, CBD, and CDSC-E is akin to that of a constant classifier, highlighting limitations in its predictive diversity. The markedly lower performance of older models like GPT-2-xl, with an average score of 13.7 points, illustrates the rapid evolution and significant enhancements in model capabilities. Llama-2-13b-chat has the lowest average of 33.7 among a few billion parameter models.  

Our study shows that in the second group of models, the ones made especially for Polish work better. Notably, Bielik-7b-instruct, a derivative of the Mistral-7B-v0.1 architecture fine-tuned on Polish instructions, demonstrates enhanced capabilities. Mistral-7b-instruct, which is fine-tuned in English is slightly worse. However, Krakowiak-7b-v2, which also derives from Mistral-7b-v0.1, does not outperform Mistral-7b-instruct. This suggests that continual fine-tuning on the target language must be approached with caution.

The second-best Polish-based model is Trurl-2-13b, which is built on the Llama-2-13b framework. A comparative analysis with Llama-2-13b-chat from the third cohort illustrates the effectiveness of fine-tuning on a Polish corpus, leading to improved overall metrics and a reduction in erroneous predictions. This underscores the significance of targeted fine-tuning in enhancing model performance in language-specific applications.

\subsubsection{Scaling for Llama}
\begin{figure}[!h]
    \centering
    \includegraphics[width=\linewidth]{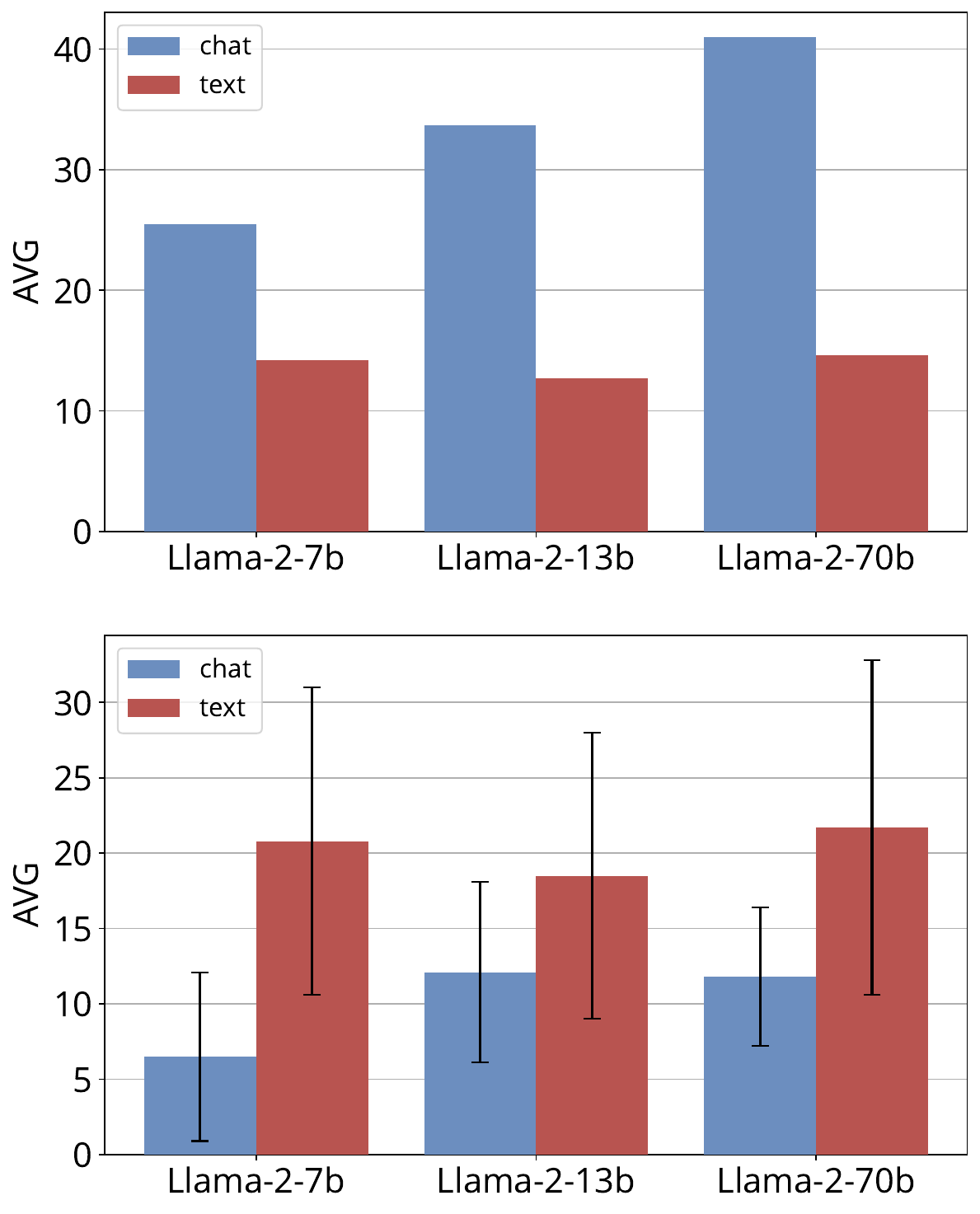}
    \caption{Average performance (AVG) for Llama-2 family models. The upper figure denotes the zero-shot performance. The lower figure denotes a few-shot performance with 16 shots. The \textit{text} model is trained with language modeling on the general corpus. The \textit{chat} are continually trained on instruction datasets.}
    \label{fig:llama-scaling}
\end{figure}

Figure~\ref{fig:llama-scaling} depicts the average performance across different scales within the Llama-2 model family, underscoring the influence of instruction fine-tuning on performance metrics. In the zero-shot paradigm, instruction fine-tuning augments performance, with larger models yielding enhanced results. Conversely, within the few-shot learning framework ($n=16$), the advantages of model scaling are not evident. Furthermore, in this few-shot context, instruction fine-tuning seems to negatively impact the outcomes. This is consistent with observations made in~\citep{ahuja2023mega, asai2023buffet}.

\section{Conclusion}
In this study, we introduced a few-shot benchmark for the Polish language, consisting of 7 different classification tasks. We conducted an empirical comparison between models trained with linear probing, fine-tuning, SetFit, and in-context learning. Our results show that ICL achieves the best performance for both 0 and 16 shots. However, there is a significant 14 pp gap between the zero-shot performance of GPT-4 and HerBERT-large trained on the entire dataset.

Our findings indicate that SetFit is the most effective fine-tuning technique, allowing the use of much smaller models compared to ICL. Nevertheless, there is a substantial 12.4 pp performance difference between SBERT-large and ChatGPT for 16 shots. The least effective method turned out to be simple fine-tuning with a non-linear classification head, while better results were achieved using linear probing on top of frozen embeddings.

Interestingly, we observed the highest variability in results among different LLMs employing ICL. Commercial models like PaLM-2, GPT-3.5, and GPT-4 show excellent comprehension of Polish and outperform open-source models. However, Mistral-7b-instruct also performs exceptionally well, despite not being specifically trained on Polish corpora. The same cannot be said for Llama-2.

Our tests confirm that continual pre-training of Mistral or Llama-2 on Polish corpora yields better zero-shot results in our benchmark. Both Trurl-13b and Bielik-7b-instruct perform better than Llama-2-13b-chat and Mistral-7b-instruct. However, in the few-shot regime, only Trurl-2-13b is better than its English counterpart. Additionally, the example of Krakowiak-7b, which was fine-tuned from Mistral-7b, demonstrates that fine-tuning must be approached with care to avoid degrading performance.

\section{Future work}
In the future, we plan to examine the generation quality of LLMs, including whether they use Polish, English, or a mixture of both in their responses. We also intend to evaluate additional open-source LLMs and integrate a wider range of tasks into our benchmark, beyond classification. Furthermore, assessing the performance of auto-prompting techniques and comparing PEFT with prompt engineering using LLMs in the few-shot regime will be of interest.

Additionally, our findings prompt intriguing inquiries. Why does Mistral-7b-instruct perform so well in Polish, even though it was not specifically trained on a large-scale Polish corpus? This contributes to the broader research on adapting pre-trained LLMs to non-English languages~\citep{zhao2024llama,zhu2023extrapolating,chen-etal-2024-monolingual,csaki2023efficiently,rostami2024persianmind,basile2023llamantino} and in-context cross-lingual transfer~\citep{etxaniz2023multilingual,zhang-etal-2023-dont,lai2023chatgpt,asai2023buffet,ahuja2023mega,bandarkar2023belebele}.
\section{Limitations}
Our goal was to develop an evaluation methodology that reflects real-world situations and defines the maximum potential of few-shot learning for tasks in the Polish language. Previous studies~\citep{perez2021true, alex2022raft} have shown that using a validation dataset for hyperparameter tuning may overestimate the capabilities of "true few-shot learning," with results varying greatly depending on the chosen seed or template.
The number of few-shot examples was quite small and unevenly distributed across tasks. We have datasets ranging from 8 examples per class (PAC, DYK) to 1 (CST-Wikinews), affecting the interpretation of results and leading to potential label biases.

Although automatic evaluation metrics are widely used and convenient for assessing models, human evaluation remains the definitive standard for certain tasks. In this study, we used two metrics influenced by automatic evaluation. For the ICL, we considered the label name as the definitive truth, ignoring synonyms or terms with similar meanings within the task. For greedy decoding, we applied exact match metrics with minimal preprocessing. Our analysis indicates that this method is unlikely to significantly affect the results, but the challenge of automatic evaluation remains, particularly with generative LLMs.

Our benchmark uses diverse, classification-only datasets to measure model performance in different areas, including datasets of varying lengths and number of classes. However, it is limited to open-source Polish datasets, introducing several challenges. For instance, the CBD dataset had noisy data and low quality, while the clarity of the CST-Wikinews dataset was questionable, even to the original authors. Additionally, the availability of datasets in Polish is much less than in English, limiting our ability to create a comprehensive benchmark.

The list of few-shot techniques in the NLP is extensive and continues to grow each year. We selectively chose a few based on ease of use, recognition, potential benefit, and computational effort required. For instance, we did not perform any text-to-text fine-tuning with LLMs, which might have yielded better results than prompt engineering. Similarly, the number of open-source LLMs suitable for ICL is increasing rapidly.

\section{Reproducibility}
Publicly available datasets were utilized to establish the benchmark, with references to these datasets delineated in the Appendix~\ref{app:datasets}. The models employed in this investigation include both open-source variants and those accessible via paid subscriptions through their respective APIs (see Appendix~\ref{app:models}). The provision of commercial models was facilitated through platforms such as Google and Azure, with specific versions of these models being fixed for the duration of the study. To ensure experimental consistency, fixed random seeds were used, and the output generation from the language models was performed at a temperature setting of 0 to maintain deterministic behavior (see Appendix~\ref{app:in-context-learning}). Additionally, the values of the hyperparameters used for fine-tuning experiments are provided in Appendix~\ref{app:hyperparameter}.

\section*{Acknowledgements}
We express our gratitude to Allegro.pl Sp. z o.o. for providing the funding for computational resources.

All content represents the opinion of the authors, which is not necessarily shared or endorsed by their respective employers.

\bibliography{custom}
\bibliographystyle{acl_natbib}

\appendix
\include{appendix}
\end{document}

%% file: appendix.tex
\onecolumn
\section{Table with the results from all experiments}
This table presents collected in one place results from all experiments mentioned in the paper. Results for baseline HerBERT-large in red color denotes results reported on KLEJ benchmark (and they are strongly differ with our experiments). For ICL zero-shot experiments subscript indicates a fraction of the test set with invalid predictions. For experiments with "±" in subscript, we use subscript to denote standard deviation.

{\small
\begin{longtable}{l|cccccccc}
\toprule
& \rot{PAC} & \rot{Polemo2} & \rot{CBD} & \rot{NKJP-NER} & \rot{DYK} & \rot{CDSC-E} & \rot{CST-Wikinews}\\
\midrule
\endfirsthead

\midrule
\multicolumn{8}{c}{Continue of previous page table} \\
\midrule
& \rot{PAC} & \rot{Polemo2} & \rot{CBD} & \rot{NKJP-NER} & \rot{DYK} & \rot{CDSC-E} & \rot{CST-Wikinews}\\
\midrule

\endhead
    \multicolumn{8}{c}{Baselines} \\
    \midrule
    XGBoost & 56.4\down{ ± 16.7} & 38.5\down{ ± 3.4} & 19.7\down{ ± 6.0} & 20.9\down{ ± 4.7} & 22.6\down{ ± 5.1} & 57.5\down{ ± 10.4} & 11.4\down{ ± 1.5} \\
    Random Guessing & 57.5\down{ ± 0.7} & 24.8\down{ ± 1.0} & 20.9\down{ ± 1.6} & 16.8\down{ ± 0.8} & 25.5\down{ ± 1.7} & 33.6\down{ ± 0.7} & 6.9\down{ ± 0.9} \\
    Constant Classifier & 80.6 & 41.3 & 23.6 & 34.3 & 28.9 & 74.4 & 16.9 \\
    HerBERT-large & 91.1\down{ ± 0.0} & 90.9\down{ ± 0.0} & 53.2\down{ ± 3.2} \textcolor{red}{(72.0)} & 94.0\down{ ± 0.0} & 68.8\down{ ± 2.1} \textcolor{red}{(75.8)} & 93.4\down{ ± 0.0} & 67.9\down{ ± 1.0} \\
    \midrule
    \multicolumn{8}{c}{Linear probing} \\
    \midrule
    SBERT-large &67.5\down{ ± 7.0} & 60.3\down{ ± 4.6} & 40.1\down{ ± 4.3} & 30.4\down{ ± 5.9} & 27.2\down{ ± 1.8} & 62.8\down{ ± 5.8} & 15.3\down{ ± 3.0} \\
    HerBERT-large & 72.6\down{ ± 10.3} & 47.0\down{ ± 3.1} & 33.8\down{ ± 10.1} & 27.8\down{ ± 3.3} & 34.6\down{ ± 2.8} & 71.0\down{ ± 6.1} & 14.2\down{ ± 2.7} \\
    RoBERTa-large & 67.8\down{ ± 11.3} & 57.9\down{ ± 4.4} & 37.4\down{ ± 6.1} & 27.6\down{ ± 3.9} & 25.1\down{ ± 3.0} & 59.2\down{ ± 7.5} & 14.5\down{ ± 2.7}  \\
    GPT-2-large & 72.0\down{ ± 2.9} & 48.3\down{ ± 4.6} & 36.4\down{ ± 4.1} & 25.2\down{ ± 4.3} & 28.6\down{ ± 3.1} & 54.4\down{ ± 7.5} & 13.6\down{ ± 3.0} \\
    Llama-2-70b\raise0.5ex\hbox{*} & 67.6\down{ ± 12.5} & 41.4\down{ ± 2.9} & 28.4\down{ ± 2.5} & 24.2\down{ ± 2.0} & 30.7\down{ ± 3.0} & 61.1\down{ ± 6.1} & 13.5\down{ ± 4.2}\\
    \midrule
    Ada & 72.9\down{ ± 6.3} & 55.2\down{ ± 5.0} & 30.7\down{ ± 4.2} & 29.1\down{ ± 2.5} & 25.4\down{ ± 4.1} & 58.2\down{ ± 11.7} & 14.8\down{ ± 3.9}\\
    DaVinci & 67.6\down{ ± 7.0} & 58.9\down{ ± 1.3} & 36.9\down{ ± 9.7} & 30.1\down{ ± 6.6} & 29.9\down{ ± 2.7} & 60.1\down{ ± 11.2} & 15.5\down{ ± 3.5}\\
    Gecko &  61.8\down{ ± 4.5} & 42.9\down{ ± 6.4} & 23.7\down{ ± 1.3} & 24.8\down{ ± 4.3} & 25.0\down{ ± 3.8} & 68.8\down{ ± 6.3} & 17.0\down{ ± 2.8}\\
    Gecko multilingual & 72.2\down{ ± 5.6} & 60.8\down{ ± 2.6} & 41.0\down{ ± 3.3} & 28.4\down{ ± 2.8} & 27.4\down{ ± 2.3} & 62.1\down{ ± 12.6} & 16.2\down{ ± 5.7}\\
    \midrule

    \multicolumn{8}{c}{Fine-tuning with non-linear head} \\
    \midrule
    SBERT-large & 33.6\down{ ± 31.3} & 47.0\down{ ± 5.8} & 32.0\down{ ± 30.7} & 28.1\down{ ± 7.7} & 6.4\down{ ± 10.3} & 61.0\down{ ± 17.6} & 10.1\down{ ± 4.3} \\
    HerBERT-large  & 26.7\down{ ± 23.4} & 29.5\down{ ± 9.7} & 30.8\down{ ± 36.6} & 25.2\down{ ± 3.8} & 16.2\down{ ± 29.7} & 44.4\down{ ± 26.7} & 8.7\down{ ± 3.7} \\
    RoBERTa-large  & 6.2\down{ ± 11.5} & 36.4\down{ ± 4.3} & 15.3\down{ ± 10.5} & 21.3\down{ ± 6.1} & 27.2\down{ ± 0.4} & 54.5\down{ ± 26.0} & 8.7\down{ ± 2.8} \\
    \midrule
    \multicolumn{8}{c}{SetFit} \\
    \midrule
    SBERT-large  & 68.8\down{ ± 6.5} & 69.9\down{ ± 10.7} & 44.4\down{ ± 5.4} & 30.7\down{ ± 6.5} & 27.7\down{ ± 2.9} & 72.3\down{ ± 5.7} & 16.2\down{ ± 3.2} \\
    HerBERT-large & 70.7\down{ ± 11.9} & 46.0\down{ ± 11.2} & 42.5\down{ ± 10.9} & 25.8\down{ ± 11.6} & 40.6\down{ ± 4.7} & 67.2\down{ ± 14.5} & 15.7\down{ ± 1.9} \\
    \midrule
    \multicolumn{8}{c}{In-context learning} \\
    \midrule       
    /ICL\down{n=0}/ GPT-4 & {83.3}\down{.0} & {82.6}\down{.009} & {60.6}\down{.028} & {58.8}\down{.004} & {81.3}\down{.002} & {76.0}\down{.0} & {18.8}\down{.003} \\
    /ICL\down{n=0}/ GPT-3.5 & 82.2\down{.014} & 81.6\down{.005} & 50.0\down{.046} & 44.9\down{.001} & 53.1\down{.0} & 62.9\down{.0} & 13.3\down{.003} \\
    /ICL\down{n=16}/ GPT-3.5 & 73.9\down{ ± 3.6} & 81.9\down{ ± 2.1} & 64.1\down{ ± 1.9} & 46.1\down{ ± 2.9} & 64.1\down{ ± 1.8} & 66.7\down{ ± 7.6} & 19.8\down{ ± 2.7} \\
    /ICL\down{n=0}/ Bison-text & 80.2\down{.006} & 80.7\down{.009} & 42.6\down{.077} & 47.5\down{.027} & 61.6\down{.016} & 35.0\down{.001} & 17.7\down{.003} \\
    /ICL\down{n=16}/ Bison-text & 80.2\down{.006} & 80.7\down{.009} & 42.6\down{.077} & 47.5\down{.027} & 61.6\down{.016} & 35.0\down{.001} & 17.7\down{.003} \\
    \midrule

    /ICL\down{n=0}/ Llama-2-7b  & 34.6\down{.659} & 12.3\down{.379} & 0.5\down{.744} & 14.0\down{.310} & 27.5\down{.014} & 6.6\down{.0} & 3.9\down{.409}\\
    /ICL\down{n=16}/ Llama-2-7b & 52.8\down{ ± 13.7} & 0.0\down{ ± 0.0} & 20.5\down{ ± 2.5} & 11.8\down{ ± 8.3} & 31.9\down{ ± 1.9} & 26.2\down{ ± 28.0} & 2.1\down{ ± 0.0} \\
    /ICL\down{n=0}/ Llama-2-13b  & 37.1\down{.606} & 8.9\down{.54} & 1.0\down{.686} & 13.2\down{.225} & 17.2\down{.136} & 6.6\down{.003} & 4.7\down{.036}\\
    /ICL\down{n=16}/ Llama-2-13b & 38.9\down{ ± 24.5} & 0.0\down{ ± 0.0} & 15.1\down{ ± 6.5} & 0.7\down{ ± 0.5} & 38.1\down{ ± 4.3} & 34.4\down{ ± 24.0} & 2.1\down{ ± 0.0}\\
    /ICL\down{n=0}/ Llama-2-70b  & 41.8\down{.576} & 11.3\down{.678} & 0.3\down{.662} & 21.4\down{.324} & 19.2\down{.055} & 6.6\down{.0} & 1.6\down{.859} \\
    /ICL\down{n=16}/ Llama-2-70b  & 58.9\down{ ± 9.1} & 0.0\down{ ± 0.0} & 20.7\down{ ± 8.6} & 21.5\down{ ± 10.0} &40.7\down{ ± 20.1} & 6.6\down{ ± 0.0} & 3.3\down{ ± 1.2} \\

    /ICL\down{n=0}/ Llama-2-7b-chat & 69.6\down{.146} & 46.7\down{.306} & 7.9\down{.601} & 11.4\down{.276} & 29.1\down{.0} & 8.9\down{.789} & 4.7\down{.388} \\
    /ICL\down{n=16}/ Llama-2-7b-chat & 30.8\down{ ± 38.2}& 0.0\down{ ± 0.0} & 0.4\down{ ± 1.0} & 1.3\down{ ± 2.7} & 12.7\down{ ± 9.8} & 0.0\down{ ± 0.0} & 0.0\down{ ± 0.0} \\
    
    /ICL\down{n=0}/ Llama-2-13b-chat & 56.1\down{.069} & 62.7\down{.048} & 10.1\down{.508} & 4.6\down{.785} & 37.3\down{.024} & 63.1\down{.029} & 1.8\down{.797} \\
    /ICL\down{n=16}/ Llama-2-13b-chat & 47.8\down{ ± 41.7} & {0.0}\down{ ± 0.0} & 0.2\down{ ± 0.4} & 3.9\down{ ± 7.7} & 20.0\down{ ± 8.8} & 12.1\down{ ± 18.9} & 1.1\down{ ± 2.4}\\
    
    /ICL\down{n=0}/ Llama-2-70b-chat & 73.4\down{.090} & 79.8\down{.002} & 30.8\down{.122} & 37.3\down{.087} & 28.9\down{.0} & 19.8\down{.0} & 17.2\down{.016} \\
    /ICL\down{n=16}/ Llama-2-70b-chat & 38.4\down{ ± 28.6} & 0.0\down{ ± 0.0} & 10.2\down{ ± 6.4} & 16.7\down{ ± 13.0} & 0.0\down{ ± 0.0} & 5.8\down{ ± 1.9} & 11.5\down{ ± 4.0} \\

    \midrule
    /ICL\down{n=0}/ Trurl-2-13b & 74.1\down{.064} & 79.5\down{.001} & 50.0\down{.001} & 33.6\down{.038} & 49.7\down{.015} & 50.2\down{.001} & 14.6\down{.083}\\
    /ICL\down{n=16}/ Trurl-2-13b & 66.7\down{ ± 12.0} & {0.0}\down{ ± 0.0} & 12.1\down{ ± 6.6} & 15.9\down{ ± 11.3} & 30.8\down{ ± 1.4} & 26.1\down{ ± 21.8} & 6.3\down{ ± 3.2} \\
    /ICL\down{n=0}/ Krakowiak-7b-v2 & {80.6}\down{.001} & 65.0\down{.018} & {23.6}\down{.0} & 16.7\down{.024} & 41.4\down{.002} & {74.4}\down{.0} & 6.0\down{.034} \\ 
    /ICL\down{n=16}/ Krakowiak-7b-v2 & 57.9\down{ ± 22.5} & 14.4\down{ ± 18.5} & 0.0\down{ ± 0.0} & 13.8\down{ ± 1.0} & 28.9\down{ ± 0.1} & 38.0\down{ ± 24.5} & 4.8\down{ ± 4.7} \\
    /ICL\down{n=0}/ Bielik-7b-instruct & 78.1\down{.034} & 77.3\down{.0} & 53.9\down{.0} & 31.0\down{.01} & 49.0\down{.001} & 65.3\down{.0} & 6.0\down{.023} \\
    /ICL\down{n=16}/ Bielik-7b-instruct & 76.1\down{ ± 4.7} & {0.0}\down{ ± 0.0} & 23.7\down{ ± 0.1} & 23.2\down{ ± 2.2} & 38.3\down{ ± 3.8} & 55.0\down{ ± 10.7} & 5.4\down{ ± 1.3} \\
    /ICL\down{n=0}/ Mistral-7b-instruct & 78.8\down{.036} & 70.0\down{.043} & 40.2\down{.001} & 36.8\down{.054} & 52.7\down{.001} & 65.2\down{.0} & 11.7\down{.078} \\
    /ICL\down{n=16}/ Mistral-7b-instruct & 48.8\down{ ± 6.8} & 73.5\down{ ± 6.1} & 34.4\down{ ± 8.6} & 28.2\down{ ± 1.8} & 45.1\down{ ± 1.6} & 77.6\down{ ± 6.5} & 17.9\down{ ± 2.4} \\
    /ICL\down{n=0}/ GPT-2-xl & 27.0\down{.619} & 5.1\down{.827} & 2.9\down{.794} & 13.3\down{.513} & 18.9\down{.3} & 29.1\down{.519} & 0.0\down{.992} \\
    /ICL\down{n=16}/ GPT-2-xl & 24.8\down{ ± 24.7} & {0.0}\down{ ± 0.0} & 0.0\down{ ± 0.0} & 14.0\down{ ± 11.2} & 16.9\down{ ± 1.8} & 74.0\down{ ± 0.5} & 5.2\down{ ± 5.7} \\
\label{tab:icl_full_few_shot}
\end{longtable}
}
\clearpage 

\section{Datasets}\label{app:datasets}
Table~\ref{tab:app_datasets} describes the source names of datasets as cataloged within the HuggingFace datasets repository.
\begin{table*}[h!]
\centering
  \begin{tabular}{c|l}
  \toprule
    Abbrev. & HuggingFace name \\
    \midrule
    PAC & \verb|allegro/abusive-clauses-pl| \\
    DYK & \verb|allegro/klej-dyk| \\
    CDSC-E & \verb|allegro/klej-cdsc-e| \\
    Polemo2 & \verb|clarin-pl/polemo2-official| \\
    CBD & \verb|allegro/klej-cbd| \\
    NKJP-NER & \verb|allegro/klej-nkjp-ner| \\
    CST-Wikinews & \verb|clarin-pl/cst-wikinews| \\
    \bottomrule
  \end{tabular}
  \caption{Abbreviation and name of the HugginFace equivalent used in this paper.}
  \label{tab:app_datasets}
\end{table*}

\section{Hyper-parameter tuning}\label{app:hyperparameter}
For each task and random seed, we search for the best hyperparameters using Optuna with Tree-structured Parzen Estimator (TPE). We set up 100 total trials and 25 warmup trials. The exception was linear probing with logistic regression where we set 20 trials and 5 warmups. We chose the macro-F1 score as the target metric for each task. A list of hyperparameters can be found in Table~\ref{tab:hyperparamaters}. For fine-tuning with non-linear head, we used a fixed batch size of $8$ and a single Nvidia V100 GPU. For SetFit we used a single Nvidia A100 80GB GPU. 

\begin{table*}[h!]
    \centering\small
    \begin{tabular}{c|l}
        \toprule
        Method & Optuna \\
        \midrule
        \multirow{5}{*}{XGBoost} & \verb|si("n_estimators", 50, 1000, 50)| \\
        & \verb|si("max_depth", 4, 15, 1)| \\
        & \verb|su("subsample", 0.5, 1.0)| \\
        & \verb|sl("learning_rate", 1.0e-07, 1.0)| \\
        & \verb|si("vocab_size", 10, 1000, 50)| \\
        \midrule
        \multirow{2}{*}{Logistic regression} & \verb|sc("solver", ["lbfgs", "liblinear", "newton-cg"])| \\
        & \verb|sf("C", 1e-05, 100)| \\
        \midrule
        \multirow{4}{*}{SetFit} & \verb|sf("learning_rate", 1e-6, 1e-3, log=True)| \\
        & \verb|si("num_epochs", 1, 5)| \\
        & \verb|sc("batch_size", [4, 8, 16, 32])| \\
        & \verb|sc("num_iterations", [5, 10, 20])| \\
        \midrule
        \multirow{3}{*}{LoRA} & \verb|si("lora_alpha", 8, 16, 4)| \\
        & \verb|si("r", 8, 16, 4)| \\
        & \verb|su("lora_dropout", 0.0, 0.5)| \\
        \midrule
        \multirow{5}{*}{Fine-tuning with non-linear head} & \verb|sc("lr", [1e-06, 5e-06, 1e-05, 5e-05, 1e-04, 5e-04, 1e-03, 3e-03])| \\
        & \verb|sc("weight_decay", [0.0, 1e-04, 1e-03, 1e-02, 1e-01])| \\
        & \verb|sc("classifier_dropout", [0.0, 0.1, 0.2, 0.3, 0.4, 0.5])| \\
        & \verb|sc("max_epochs", [2, 5, 10, 20, 50, 70, 100])| \\
        & \verb|sc("eps", [1e-08, 1e-07, 1e-06, 1e-05, 1e-04])| \\
        \bottomrule
    \end{tabular}
    \caption{List of hyperparameters for each training method using Optuna Python notation. We use short-hand notation: \texttt{suggest\_int (si)}, \texttt{suggest\_uniform (su)}, \texttt{suggest\_loguniform (sl)}, \texttt{suggest\_categorical (sc)}, \texttt{suggest\_float (sf)}.}
    \label{tab:hyperparamaters}
\end{table*}

\section{Pretrained models}\label{app:models}
Table~\ref{tab:models-info} summarizes the model naming convention we used in the paper and their corresponding versions and providers. Commercial models from Google and Microsoft were accessed via API.
\begin{table*}[h!]
    \centering\small
    \begin{tabular}{c|c|c|c|c}
        \toprule
        Model & Version & Provider & Language & Context\\
        \midrule
         SBERT-large & \verb|Voicelab/sbert-large-cased-pl| & HF & PL & 512 \\
         HerBERT-large & \verb|allegro/herbert-large-cased| & HF & PL & 512 \\
         RoBERTa-large & \verb|sdadas/polish-roberta-large-v2| & HF & PL & 512 \\
         GPT-2-xl & \verb|sdadas/polish-gpt2-xl| & HF & PL & 1536 \\
         Krakowiak-v2-7b & \verb|szymonrucinski/krakowiak-v2-7b| & HF & PL & 32768 \\
         Trurl-13b & \verb|Voicelab/trurl-2-13b| & HF & PL & 4096 \\
         Mistral-7b-instruct & \verb|mistralai/Mistral-7B-Instruct-v0.2| & HF & EN & 32768 \\
         Bielik-7b-instruct & \verb|speakleash/Bielik-7B-Instruct-v0.1| & HF & PL & 4096 \\
         Llama-2-7b & \verb|Llama 2 7b| & Meta & EN & 4096 \\
         Llama-2-13b & \verb|Llama 2 13b| & Meta & EN & 4096 \\
         Llama-2-70b & \verb|Llama 2 70b| & Meta & EN & 4096 \\
         Llama-2-7b-chat & \verb|Llama 2 7b-chat| & Meta & EN & 4096 \\
         Llama-2-13b-chat & \verb|Llama 2 13b-chat| & Meta & EN & 4096 \\
         Llama-2-70b-chat & \verb|Llama 2 70b-chat| & Meta & EN & 4096 \\
         Ada & \verb|text-embedding-ada-002| & MS Azure & EN & 8191 \\
         DaVinci & \verb|text-search-davinci-doc-001| & MS Azure & EN & - \\
         GPT-3.5 & \verb|gpt-35-turbo (version=0301)| & MS Azure & Unknown & 18384  \\
         GPT-4 & \verb|gpt-4 (version=0613)| & MS Azure & Unknown & 8192 \\
         Gecko & \verb|textembedding-gecko@001| & Google & EN & - \\
         Gecko Multilingual & \verb|textembedding-gecko-multilingual@001| & Google & Multi & - \\
         Bison-text & \verb|text-bison@001| & Google & Unknown & 8192 \\
         \midrule\midrule
         \multicolumn{4}{c}{Not covered in the paper} \\
         \midrule\midrule
         Trurl-7b & \verb|Voicelab/trurl-2-7b| & HF & PL & 4096 \\
         Krakowiak-7b & \verb|szymonrucinski/krakowiak-7b| & HF & PL & 2048 \\
         mT0-\{small,base,\\large,xl,xxl\} & \verb|bigscience/mt0-{x}| & HF & Multi & $\infty$ \\
         XGLM-\{564M,1.7B,\\2.9B,4.5B,7.5B\} & \verb|facebook/xglm-{x}| & HF & Multi & 2048 \\
         Gemma-\{2b,7b\}-it & \verb|google/gemma-{x}-it| & HF & EN & 8192 \\
         OpenChat & \verb|openchat/openchat-3.5-0106| & HF & EN & 8192 \\
         Mixtral & \verb|mistralai/Mixtral-8x7B-Instruct-v0.1| & HF & EN,FR,IT,ES,DE & 32768 \\
         Starling & \verb|berkeley-nest/Starling-LM-7B-alpha| & HF & EN & 8192 \\
         DBRX & \verb|databricks/dbrx-instruct| & HF & EN & 32768\\
         Text-DaVinci & \verb|text-davinci-003| & MS Azure & EN & - \\
         Bison-chat & \verb|chat-bison@001| & Google & Unknown & 8192\\
         \bottomrule
    \end{tabular}
    \caption{Information on pre-trained models used in the paper or left for future exploration. The penultimate column indicates which language was used for final fine-tuning. We are not sure how commercial models were pre-trained so we left \textit{Unknown} indicator. Abbreviations: HF - HuggingFace, MS - Microsoft.}
    \label{tab:models-info}
\end{table*}
\section{In-context learning}\label{app:in-context-learning}
Each LLM under evaluation was configured to operate deterministically with parameters set to \verb|temperature=0| and \verb|max_tokens=20|, while other parameters remained at their default values. Access to GPT-3.5 predictions was facilitated through the \texttt{guidance}~\footnote{\url{https://github.com/guidance-ai/guidance/tree/0.0.64}} library, and PaLM 2 predictions were obtained via the \texttt{Vertex AI API}~\footnote{\url{https://github.com/googleapis/python-aiplatform/tree/v1.29.0}}. The official repository \texttt{llama}~\footnote{\url{https://github.com/facebookresearch/llama}} was utilized for interfacing with Llama-2, which had its context length parameter fixed at 4096 tokens. In comparison, GPT-3.5 and Bison-text were configured with context lengths of 16,384~\footnote{\url{https://platform.openai.com/docs/models/gpt-3-5}} and 8,192~\footnote{\url{https://cloud.google.com/vertex-ai/docs/generative-ai/learn/models}} tokens, respectively. Inference for other open-source models was conducted using the \texttt{vLLM} library~\cite{kwon2023efficient}.

Regarding computational resources, the Llama-2-7B model execution was supported by a single A100 40GB GPU. The Llama-2-13B model required 4 L4 GPUs, whereas the Llama-2-70B model utilized 8 L4 GPUs, demonstrating the scalable nature of the hardware deployment in correspondence with the increasing model size and computational demands.

\newpage
\section{Sample prompts}
In the following, we illustrate a sample few-shot ($n=2$) prompt utilized for the NKJP-NER task, employing models such as GPT-4/GPT-3.5, Llama-chat, and Mistral-instruct. Each model employs special characters to delineate roles within the prompt. The Krakowiak-v2 and Trurl are based on Mistral and Llama respectively.

\begin{figure}[h!]
\small\begin{lstlisting}[
    language=gpt, 
    frame=single
]
<|im_start|>system
Rozwiązujesz zadanie klasyfikacji dla języka polskiego.<|im_end|>
<|im_start|>user
Mamy kilka encji: geografia, brak, organizacja, imię, miejsce, czas.
Zdanie "Ma do tego dojść na specjalnym posiedzeniu Rady Miejskiej , prawdopodobnie zamkniętym dla gości i mediów ." zawiera encję organizacja
Zdanie "W dole widać i słychać aleje Stryjskiego parku , rąbek stawu błyszczy , drzewa na przeciwległym wzgórzu czerwienią się od zachodu ." zawiera encję geografia
Zdanie "Przychodnia dotkliwie odczuła więc jego brak ." zawiera encję <|im_end|>
<|im_start|>assistant
\end{lstlisting}

\small\begin{lstlisting}[
    language=gpt, 
    frame=single
]
[INST] <<SYS>>
Rozwiązujesz zadanie klasyfikacji dla języka polskiego.
<</SYS>>

Mamy kilka encji: geografia, brak, organizacja, imię, miejsce, czas.
Zdanie "Ma do tego dojść na specjalnym posiedzeniu Rady Miejskiej , prawdopodobnie zamkniętym dla gości i mediów ." zawiera encję organizacja
Zdanie "W dole widać i słychać aleje Stryjskiego parku , rąbek stawu błyszczy , drzewa na przeciwległym wzgórzu czerwienią się od zachodu ." zawiera encję geografia
Zdanie "Przychodnia dotkliwie odczuła więc jego brak ." zawiera encję [/INST]
\end{lstlisting}

\small\begin{lstlisting}[
    language=gpt, 
    frame=single
]
<s>[INST] Rozwiązujesz zadanie klasyfikacji dla języka polskiego.
Mamy kilka encji: geografia, brak, organizacja, imię, miejsce, czas.
Zdanie "Ma do tego dojść na specjalnym posiedzeniu Rady Miejskiej , prawdopodobnie zamkniętym dla gości i mediów ." zawiera encję organizacja
Zdanie "W dole widać i słychać aleje Stryjskiego parku , rąbek stawu błyszczy , drzewa na przeciwległym wzgórzu czerwienią się od zachodu ." zawiera encję geografia
Zdanie "Przychodnia dotkliwie odczuła więc jego brak ." zawiera encję [/INST]
\end{lstlisting}

\caption{Sample prompt with two examples for the GPT-4/GPT-3.5 models (top), Llama-2-chat (middle) and Mistral-instruct (bottom).}
\end{figure}

\section{AI assistant}
We used the GPT-4 model from OpenaAI to improve the language quality of this manuscript.

\section{Prompt templates}\label{app:templates}
\subsection{PAC}
\input{templates/pac_templates}
\subsection{DYK}
\input{templates/dyk_templates}
\subsection{CDSC-E}
\input{templates/cdsce_templates}
\subsection{Polemo2}
\input{templates/polemo2_templates}
\subsection{CBD}
\input{templates/cbd_templates}
\subsection{NKJP-NER}
\input{templates/nkjp_ner_templates}
\subsection{CST-Wikinews}
\input{templates/cst_wikinews_templates}

%% file: templates/pac_templates.tex
\small\begin{minted}[breaklines]{yaml}
f535ad28-8896-43a3-aa4c-f41c296848de:
  instruction: 'Czy następująca klauzula jest uczciwa?'
  demonstration: '{text} Ocena: {labels}'
  label_mapping:
    0: Nie
    1: Tak
\end{minted}
\small\begin{minted}[breaklines]{yaml}
a68bdcea-0c7f-42a8-8987-4d7114b43658:
  instruction: ''
  demonstration: '{text}. Fragment tekstu jest {labels}'
  label_mapping:
    0: klauzulą abuzywną
    1: bezpiecznym postanowieniem umowy
\end{minted}
\small\begin{minted}[breaklines]{yaml}
e3fb857a-7bd9-4268-8915-c7f494374ca6:
  instruction: ''
  demonstration: 'Klauzula: {text} Ocena: {labels}'
  label_mapping:
    0: niedozwolona
    1: dozwolona
\end{minted}
\small\begin{minted}[breaklines]{yaml}
35a1ac4a-c0a9-4138-ab0d-cd759cdc805f:
  instruction: 'Oceń czy następujący zapis jest uczciwy lub nieuczciwy.'
  demonstration: '{text} => {labels}'
  label_mapping:
    0: nieuczciwy
    1: uczciwy
\end{minted}
\small\begin{minted}[breaklines]{yaml}
735df98d-2732-4d82-9b49-39de04a435e8:
  instruction: 'Do jakiej klasy przypisałbyś następującą klauzulę?

    Klasy:

    - uczciwa

    - nieuczciwa'
  demonstration: '{text} => {labels}'
  label_mapping:
    0: nieuczciwa
    1: uczciwa
\end{minted}
\small\begin{minted}[breaklines]{yaml}
5d80ad77-791b-4acc-9676-84d2d3d94966:
  instruction: 'Klauzule abuzywne to postanowienia w umowach konsumenckich, które nie są
    uzgodnione indywidualnie z klientem. Czy fragment jest klauzulą abuzywną?'
  demonstration: '{text} => {labels}'
  label_mapping:
    0: Tak
    1: Nie
\end{minted}
\small\begin{minted}[breaklines]{yaml}
e9ef6dcc-e21e-4efd-827e-d7cd35a97ce4:
  instruction: 'Klauzule abuzywne to postanowienia w umowach konsumenckich, które nie są
    uzgodnione indywidualnie z klientem. Czy fragment jest klauzulą abuzywną? Tak lub
    Nie?'
  demonstration: '{text} => {labels}'
  label_mapping:
    0: Tak
    1: Nie
\end{minted}
\small\begin{minted}[breaklines]{yaml}
39855489-6f18-49f3-92a8-1cdcccf0cc67:
  instruction: 'Wybierz czy klauzula jest "uczciwa" lub "nieuczciwa"'
  demonstration: '{text} => {labels}'
  label_mapping:
    0: nieuczciwa
    1: uczciwa
\end{minted}
\small\begin{minted}[breaklines]{yaml}
ceb5e55e-cb6f-47ab-b4f0-059192837c26:
  instruction: ''
  demonstration: 'Następujący zapis {text} w umowie konsumenckiej jest {labels}'
  label_mapping:
    0: nieuczciwy
    1: uczciwy
\end{minted}
\small\begin{minted}[breaklines]{yaml}
3ad789e3-6a21-4db9-8fba-c051c37cf449:
  instruction: 'Dokończ zdanie: "Następujący zapis w umowie konsumenckiej jest".'
  demonstration: 'Zapis: {text} {labels}'
  label_mapping:
    0: nieuczciwy
    1: uczciwy
\end{minted}
\small\begin{minted}[breaklines]{yaml}
871270fa-33aa-4677-b6ce-1869632a694c:
  instruction: 'Dokończ zdanie.'
  demonstration: 'Zapis {text} to {labels}'
  label_mapping:
    0: klauzula abuzywna
    1: bezpieczne postanowienie umowne
\end{minted}
\small\begin{minted}[breaklines]{yaml}
df724698-89eb-4634-8d62-a04406b8da26:
  instruction: 'Czy tekst jest bezpiecznym postanowieniem umowy?'
  demonstration: '{text} => {labels}'
  label_mapping:
    0: Nie
    1: Tak
\end{minted}
\small\begin{minted}[breaklines]{yaml}
559e1bca-12e4-4d42-a305-60de9010cefc:
  instruction: 'Czy tekst jest bezpiecznym postanowieniem umowy? Tak lub Nie?'
  demonstration: '{text} => {labels}'
  label_mapping:
    0: Nie
    1: Tak
\end{minted}

%% file: templates/dyk_templates.tex
\small\begin{minted}[breaklines]{yaml}
8102fdda-4584-4f50-8eab-ca7e6904c70e:
  instruction: ''
  demonstration: 'Pytanie: {sentence_1} Odpowiedź: {sentence_2} Czy to prawda? {labels}'
  label_mapping:
    0: Nie
    1: Tak
\end{minted}
\small\begin{minted}[breaklines]{yaml}
37294e9a-d241-4f37-b680-df58dc625ee8:
  instruction: ''
  demonstration: 'Pytanie: {sentence_1} Odpowiedź: {sentence_2} Poprawny: {labels}'
  label_mapping:
    0: Nie
    1: Tak
\end{minted}
\small\begin{minted}[breaklines]{yaml}
5cc8dd02-287a-422b-8c1c-9583fcb5af1b:
  instruction: 'Oceń czy sugerowana odpowiedź pasuje do postawionego pytania.'
  demonstration: 'Pytanie: {sentence_1} Odpowiedź: {sentence_2} Ocena: {labels}'
  label_mapping:
    0: Nie
    1: Tak
\end{minted}
\small\begin{minted}[breaklines]{yaml}
1e8316a2-4611-40dd-9877-25c09bb4f895:
  instruction: 'Oceń czy sugerowana odpowiedź pasuje do postawionego pytania.'
  demonstration: 'Pytanie: {sentence_1} Odpowiedź: {sentence_2} Ocena: {labels}'
  label_mapping:
    0: nie pasuje
    1: pasuje
\end{minted}
\small\begin{minted}[breaklines]{yaml}
0030485b-d576-4533-aa04-8840744e13c2:
  instruction: 'Oceń czy sugerowana odpowiedź pasuje do postawionego pytania.'
  demonstration: 'Pytanie: {sentence_1} Odpowiedź: {sentence_2} Ocena: {labels}'
  label_mapping:
    0: '0'
    1: '1'
\end{minted}
\small\begin{minted}[breaklines]{yaml}
ddee4860-8329-40e5-af83-7972e17e6715:
  instruction: ''
  demonstration: 'Czy tekst: {sentence_2} odpowiada na pytanie: {sentence_1}.
    Tak lub Nie? {labels}'
  label_mapping:
    0: Nie
    1: Tak
\end{minted}
\small\begin{minted}[breaklines]{yaml}
05838165-f7b0-44bf-876d-4b3431c11dfc:
  instruction: 'Czy sugerowana odpowiedź poprawnie odpowiada na pytanie? Tak lub Nie?'
  demonstration: 'Pytanie: {sentence_1} Odpowiedź: {sentence_2} {labels}'
  label_mapping:
    0: Nie
    1: Tak
\end{minted}
\small\begin{minted}[breaklines]{yaml}
f52e97a6-546f-4f72-b026-d6ccff0fd857:
  instruction: 'Zadanie polega na przypisaniu jednej z dwóch etykiet "Tak" lub "Nie"
    do pary pytanie, odpowiedź.'
  demonstration: 'Pytanie: {sentence_1} Odpowiedź: {sentence_2} {labels}'
  label_mapping:
    0: Nie
    1: Tak
\end{minted}
\small\begin{minted}[breaklines]{yaml}
489d39ba-d7a6-4c3f-9529-caba81bea564:
  instruction: 'Mając pytanie oceń czy odpowiedź jest "prawdziwa" lub "fałszywa".'
  demonstration: 'Pytanie: {sentence_1} Odpowiedź: {sentence_2} {labels}'
  label_mapping:
    0: fałszywa
    1: prawdziwa
\end{minted}
\small\begin{minted}[breaklines]{yaml}
d78aafd2-848c-45a5-b227-09b4f0f2c000:
  instruction: 'Proszę przeczytać poniższe pytanie oraz odpowiedź i zdecydować, czy podana
    odpowiedź jest prawdziwa (P) czy fałszywa (F).'
  demonstration: 'Pytanie: {sentence_1} Odpowiedź: {sentence_2} {labels}'
  label_mapping:
    0: F
    1: P
\end{minted}
\small\begin{minted}[breaklines]{yaml}
723df916-959c-4765-936f-684f9186e161:
  instruction: 'Proszę przeczytać poniższe pytanie oraz odpowiedź i zdecydować, czy
  podana odpowiedź jest prawdziwa (P) czy fałszywa (F).'
  demonstration: 'Pytanie: {sentence_1} Odpowiedź: {sentence_2} {labels}'
  label_mapping:
    0: fałszywa
    1: prawdziwa
\end{minted}
\small\begin{minted}[breaklines]{yaml}
b3626557-c0d0-48b9-bdd0-21cb6ebc1205:
  instruction: 'Przeanalizuj krok po kroku czy podany tekst jest poprawną odpowiedzią
    na pytanie. Na końcu zwróć "tak" lub "nie".'
  demonstration: 'Pytanie: {sentence_1} Tekst: {sentence_2} {labels}'
  label_mapping:
    0: Nie
    1: Tak
\end{minted}
\small\begin{minted}[breaklines]{yaml}
4c7c5c30-61eb-49ff-89a1-8faa8e2c4c20:
  instruction: ''
  demonstration: 'Czy następujący kawałek tekstu: {sentence_2} zawiera w sobie odpowiedź
    na pytanie: {sentence_1}? Tak lub Nie? {labels}'
  label_mapping:
    0: Nie
    1: Tak
\end{minted}
\small\begin{minted}[breaklines]{yaml}
a963d2e0-fce6-4d06-ad38-e28f176bc688:
  instruction: ''
  demonstration: 'Jeśli {sentence_1} jest pytaniem to czy {sentence_2} jest pasującą
    odpowiedzią? {labels}'
  label_mapping:
    0: Nie
    1: Tak
\end{minted}

%% file: templates/cdsce_templates.tex
\small\begin{minted}[breaklines]{yaml}
063792d7-e23f-4111-a64d-b8b59057ed80:
  instruction: ''
  demonstration: 'Zdanie "{sentence_1}" odnosi się do zdania "{sentence_2}" jako {labels}'
  label_mapping:
    0: sprzeczne
    1: zgodne
    2: neutralne
\end{minted}
\small\begin{minted}[breaklines]{yaml}
7a650332-ec38-4811-8401-5889df51c80f:
  instruction: ''
  demonstration: 'Relacja "{sentence_1}" do zdania "{sentence_2}" jest {labels}'
  label_mapping:
    0: sprzeczna
    1: zgodna
    2: neutralna
\end{minted}
\small\begin{minted}[breaklines]{yaml}
05e4cd72-59a1-44fd-9d33-2a655aacc56f:
  instruction: 'Która etykieta opisuje relację między zdaniami.

    Etykiety:

    - neutralna

    - zgodna

    - sprzeczna'
  demonstration: 'Zdania: {sentence_1}; {sentence_2} Etykieta: {labels}'
  label_mapping:
    0: sprzeczna
    1: zgodna
    2: neutralna
\end{minted}
\small\begin{minted}[breaklines]{yaml}
7c5cd177-5e97-4efc-805f-86d0b5e1e6f0:
  instruction: 'Czy pierwsze zdanie pociąga za sobą drugie zdanie? Tak, Nie lub Może?'
  demonstration: 'Zdanie1: {sentence_1}

   Zdanie2: {sentence_2}

   {labels}'
  label_mapping:
    0: Nie
    1: Tak
    2: Może
\end{minted}
\small\begin{minted}[breaklines]{yaml}
a97257d8-db60-4fbe-bdc9-c198fea17706:
  instruction: ''
  demonstration: 'Załóżmy, że {sentence_1}. Czy możemy wywnioskować iż {sentence_2}? Tak,
    Nie lub Może? {labels}'
  label_mapping:
    0: Nie
    1: Tak
    2: Może
\end{minted}
\small\begin{minted}[breaklines]{yaml}
5d6b1071-9e49-422a-b649-6e04ef6dae65:
  instruction: ''
  demonstration: '{sentence_1}. Czy można powiedzieć że {sentence_2}? Tak, Nie lub Może?
    {labels}'
  label_mapping:
    0: Nie
    1: Tak
    2: Może
\end{minted}
\small\begin{minted}[breaklines]{yaml}
6ffe622f-cef2-4798-ae3e-dbfdabb0d384:
  instruction: 'Mając kontekst odpowiedz na pytanie zdanie to prawda, fałsz lub żadne.'
  demonstration: 'Kontekst: {sentence_1}

    Zdanie: {sentence_2}

    Odpowiedź: {labels}'
  label_mapping:
    0: Fałsz
    1: Prawda
    2: Żadne
\end{minted}
\small\begin{minted}[breaklines]{yaml}
f397d997-4f5f-4e08-8ce4-356f9905ea4d:
  instruction: 'Logiczne wynikanie.'
  demonstration: 'Zdanie1: {sentence_1}

    Zdanie2: {sentence_2}

    {labels}'
  label_mapping:
    0: sprzeczne
    1: zgodne
    2: neutralne
\end{minted}

%% file: templates/polemo2_templates.tex
\small\begin{minted}[breaklines]{yaml}
5882074e-1e5c-479e-b138-abdeab4e4996:
  instruction: ''
  demonstration: 'Recenzja {text} jest {labels}'
  label_mapping:
    0: neutralna
    1: negatywna
    2: pozytywna
    3: niejednoznaczna
\end{minted}
\small\begin{minted}[breaklines]{yaml}
18d75cc9-c3c0-462e-a4c5-d35bee7e810b:
  instruction: ''
  demonstration: 'Tekst {text} ma sentyment {labels}'
  label_mapping:
    0: neutralny
    1: negatywny
    2: pozytywny
    3: niejednoznaczny
\end{minted}
\small\begin{minted}[breaklines]{yaml}
a034ba19-d751-4b02-a0a2-f98e4a9d43d7:
  instruction: 'Oceń sentyment tego tekstu, używając jednej z czterech etykiet:
    pozytywny, negatywny, neutralny, ambiwalentny.'
  demonstration: 'Tekst do oceny: {text} Ocena: {labels}'
  label_mapping:
    0: neutralny
    1: negatywny
    2: pozytywny
    3: ambiwalentny
\end{minted}
\small\begin{minted}[breaklines]{yaml}
18576914-803b-4c87-812c-240a59596e37:
  instruction: 'Czy wydźwięk następującego tekstu jest pozytywny, negatywny, neutralny,
    ambiwalentny?'
  demonstration: '{text} => {labels}'
  label_mapping:
    0: neutralny
    1: negatywny
    2: pozytywny
    3: ambiwalentny
\end{minted}
\small\begin{minted}[breaklines]{yaml}
bc7b3799-4e1d-4375-a6c0-5d0e52be5750:
  instruction: 'Która z etykiet najlepiej określa sentyment tekstu?

    Etykiety:

    - neutralny

    - negatywny

    - pozytywny

    - ambiwalentny'
  demonstration: 'Tekst: {text} Etykieta: {labels}'
  label_mapping:
    0: neutralny
    1: negatywny
    2: pozytywny
    3: ambiwalentny
\end{minted}
\small\begin{minted}[breaklines]{yaml}
80c3a8cc-fe0b-441f-bd1f-abdcfc48d870:
  instruction: 'Czy to zdanie było neutralne, pozytywne, negatywne lub ambiwalentne?'
  demonstration: '{text}

    Odpowiedź: {labels}'
  label_mapping:
    0: neutralne
    1: negatywne
    2: pozytywne
    3: ambiwalentne
\end{minted}
\small\begin{minted}[breaklines]{yaml}
c49953e5-f4cf-4a86-a6c2-65d85db7f478:
  instruction: 'Mając opinię stwierdź czy jest ona neutralna, negatywna, pozytywna lub ambiwalentna?'
  demonstration: 'Opinia: "{text}"

    {labels}'
  label_mapping:
    0: neutralna
    1: negatywna
    2: pozytywna
    3: ambiwalentna
\end{minted}
\small\begin{minted}[breaklines]{yaml}
1e5de596-8d20-48fc-8615-2052a5559a77:
  instruction: 'Jaki wydźwięk ma recenzja?'
  demonstration: 'Recenzja: "{text}"

  {labels}'
  label_mapping:
    0: neutralny
    1: negatywny
    2: pozytywny
    3: ambiwalentny
\end{minted}
\small\begin{minted}[breaklines]{yaml}
a71195d9-f51f-408d-b25b-ea53dbd64421:
  instruction: 'Jaki wydźwięk ma recenzja? neutralny, negatywny, pozytywny
    lub ambiwalentny?'
  demonstration: '{text} => {labels}'
  label_mapping:
    0: neutralny
    1: negatywny
    2: pozytywny
    3: ambiwalentny
\end{minted}
\small\begin{minted}[breaklines]{yaml}
cf1247a0-463c-4ca6-b8e8-a1c8864f50e9:
  instruction: 'Przypisz sentyment do zdania.'
  demonstration: '{text} => {labels}'
  label_mapping:
    0: neutralny
    1: negatywny
    2: pozytywny
    3: ambiwalentny
\end{minted}
\small\begin{minted}[breaklines]{yaml}
173978fb-8da7-4931-b9e6-2ea1682386d6:
  instruction: 'Przypisz sentyment do zdania. Wybierz jedną z etykiet: neutralny,
    pozytywny, negatywny, ambiwalentny.'
  demonstration: '{text} => {labels}'
  label_mapping:
    0: neutralny
    1: negatywny
    2: pozytywny
    3: ambiwalentny
\end{minted}

%% file: templates/cbd_templates.tex
\small\begin{minted}[breaklines]{yaml}
3deddcad-cf5f-4efb-a1ac-8302a327a200:
  instruction: 'Czy zdanie jest obraźliwe?'
  demonstration: '{text} => {labels}'
  label_mapping:
    0: Nie
    1: Tak
\end{minted}
\small\begin{minted}[breaklines]{yaml}
23b47eed-5a7d-4388-82a0-d832c285e159:
  instruction: ''
  demonstration: 'Tekst: {text} Etykieta: {labels}'
  label_mapping:
    0: nie obraźliwe
    1: obraźliwe
\end{minted}
\small\begin{minted}[breaklines]{yaml}
62f33ae1-a7bf-4ab3-9192-afe3b5bc8e20:
  instruction: 'Czy zdanie jest obraźliwe?'
  demonstration: 'Tekst: {text} Etykieta: {labels}'
  label_mapping:
    0: nie obraźliwe
    1: obraźliwe
\end{minted}
\small\begin{minted}[breaklines]{yaml}
f1cc2961-3a7b-4cc6-887e-c661089dd994:
  instruction: 'Czy zdanie jest agresywne?'
  demonstration: '{text} => {labels}'
  label_mapping:
    0: Nie
    1: Tak
\end{minted}
\small\begin{minted}[breaklines]{yaml}
0d817d40-b514-4a40-b76a-dbdd5902bc3e:
  instruction: 'Czy zdanie jest agresywne? Tak lub Nie?'
  demonstration: '{text} => {labels}'
  label_mapping:
    0: Nie
    1: Tak
\end{minted}
\small\begin{minted}[breaklines]{yaml}
0b170e50-aec1-4843-98d6-6f51f01cc951:
  instruction: 'Czy tekst może być zaklasyfikowany jako cyberprzemoc?'
  demonstration: '{text} => {labels}'
  label_mapping:
    0: Nie
    1: Tak
\end{minted}
\small\begin{minted}[breaklines]{yaml}
83519157-5075-402c-815f-17f575f1fc5b:
  instruction: 'Czy tekst może być zaklasyfikowany jako cyberprzemoc? Tak lub Nie?'
  demonstration: '{text} => {labels}'
  label_mapping:
    0: Nie
    1: Tak
\end{minted}
\small\begin{minted}[breaklines]{yaml}
79929bb1-15a1-4232-b6da-e2a72c039446:
  instruction: 'Cyberprzemoc to stosowanie przemocy z wykorzystaniem internetu. Czy
    tekst to cyberprzemoc?'
  demonstration: '{text} => {labels}'
  label_mapping:
    0: Nie
    1: Tak
\end{minted}
\small\begin{minted}[breaklines]{yaml}
6ff460ac-1b85-4e60-b186-1360b0fca5a2:
  instruction: 'Wybierz jedną z etykiet pasującą do tekstu.

    Etykiety:

    - obraźliwe

    - nie obraźliwe'
  demonstration: 'Tekst: {text} {labels}'
  label_mapping:
    0: nie obraźliwe
    1: obraźliwe
\end{minted}

%% file: templates/nkjp_ner_templates.tex
\small\begin{minted}[breaklines]{yaml}
0fbf54ee-bfdd-4a60-866b-8c3de8ac42d6:
  instruction: 'Mamy kilka encji: geografia, brak, organizacja, imię, miejsce, czas.'
  demonstration: 'Zdanie "{text}" zawiera encję {labels}'
  label_mapping:
    1: brak
    3: imię
    4: miejsce
    0: geografia
    2: organizacja
    5: czas
\end{minted}
\small\begin{minted}[breaklines]{yaml}
00486b42-3f72-481f-a350-2b60505d297e:
  instruction: ''
  demonstration: '{text}. To zdanie dotyczy {labels}'
  label_mapping:
    1: niczego
    3: osoby
    4: miejsca
    0: geografii
    2: organizacji
    5: czasu
\end{minted}
\small\begin{minted}[breaklines]{yaml}
46c1cd05-2f1d-4de3-8de6-0f689a09d56a:
  instruction: 'Zdanie zawiera fragment odnoszący się do encji.'
  demonstration: '"{text}"

  {labels}'
  label_mapping:
    1: brak
    3: osoba
    4: miejsce
    0: geografia
    2: organizacja
    5: czas
\end{minted}
\small\begin{minted}[breaklines]{yaml}
abba041c-81a3-4392-bb02-abb28a8aa549:
  instruction: 'Którą z encji przypisałbyś zdaniu: brak, osoba, miejsce, geografia,
    organizacja, lub czas?'
  demonstration: '{text}.

    Odpowiedź: {labels}'
  label_mapping:
    1: brak
    3: osoba
    4: miejsce
    0: geografia
    2: organizacja
    5: czas
\end{minted}
\small\begin{minted}[breaklines]{yaml}
d5a15d7b-00b6-4533-ac5c-b9edaea9d484:
  instruction: 'Mamy zdanie, które może zawierać jednostkę referencyjną (encję).
    Która z nich pasuje najlepiej?'
  demonstration: '{text} => {labels}'
  label_mapping:
    1: żadna
    3: osoba
    4: miejsce
    0: geografia
    2: organizacja
    5: czas
\end{minted}
\small\begin{minted}[breaklines]{yaml}
657775d6-6547-419f-88b4-820373a3812a:
  instruction: 'Mamy zdanie, które może zawierać jednostkę referencyjną (encję).
    Która z nich pasuje najlepiej? osoba, miejsce, geografia, organizacja,
    czas lub żadna?'
  demonstration: '{text} => {labels}'
  label_mapping:
    1: żadna
    3: osoba
    4: miejsce
    0: geografia
    2: organizacja
    5: czas
\end{minted}
\small\begin{minted}[breaklines]{yaml}
b28326b4-d8e3-4a1e-845f-cdb031601873:
  instruction: 'W zdaniu może pojawić się referencja do osoby, miejsca, lokalizacji geograficznej,
    organizacji lub czasu. Która z nich lub żadna?'
  demonstration: '{text} => {labels}'
  label_mapping:
    1: żadna
    3: osoba
    4: miejsce
    0: lokalizacja geograficzna
    2: organizacja
    5: czas
\end{minted}
\small\begin{minted}[breaklines]{yaml}
ea5330fa-278e-4d87-ae69-7a427f546b7f:
  instruction: 'Zadanie polega na identyfikacji encji (lub brak) w tekście.

    Encje:

    - osoba

    - miejsce

    - geografia

    - organizacja

    - czas

    - brak'
  demonstration: 'Tekst: {text} Encja: {labels}'
  label_mapping:
    1: brak
    3: osoba
    4: miejsce
    0: geografia
    2: organizacja
    5: czas
\end{minted}
\small\begin{minted}[breaklines]{yaml}
9bd0f669-c589-4419-9cbb-773d72df8dcd:
  instruction: 'Przytoczony fragment zawiera odniesienie do encji.'
  demonstration: '"{text}".

  {labels}'
  label_mapping:
    1: niczego
    3: osoby
    4: miejsca
    0: geografii
    2: organizacji
    5: czasu
\end{minted}

%% file: templates/cst_wikinews_templates.tex
\small\begin{minted}[breaklines]{yaml}
10876076-d6a2-4ec3-b82c-f59cf79c17fa:
  instruction: ''
  demonstration: 'Zdania {sentence_1} oraz {sentence_2} to {labels}'
  label_mapping:
    0: brak relacji
    1: dalsze informacje
    2: krzyżowanie się
    3: opis
    4: parafraza
    5: spełnienie
    6: streszczenie
    7: tożsamość
    8: tło historyczne
    9: uszczegółowienie
    10: zawieranie
    11: źródło
\end{minted}
\small\begin{minted}[breaklines]{yaml}
8a0292ef-d678-4be6-bb95-0fe1b0cf136f:
  instruction: ''
  demonstration: 'Zdanie z Wikipedii "{sentence_1}" odnosi się do zdania
    "{sentence_2}" jako {labels}'
  label_mapping:
    0: brak relacji
    1: dalsze informacje
    2: krzyżowanie się
    3: opis
    4: parafraza
    5: spełnienie
    6: streszczenie
    7: tożsamość
    8: tło historyczne
    9: uszczegółowienie
    10: zawieranie
    11: źródło
\end{minted}
\small\begin{minted}[breaklines]{yaml}
f2fd82fd-f3d9-4de0-aeb5-1801d2b5128b:
  instruction: ''
  demonstration: 'Zdanie_1: {sentence_1} Zdanie_2: {sentence_2} Etykieta: {labels}'
  label_mapping:
    0: brak relacji
    1: dalsze informacje
    2: krzyżowanie się
    3: opis
    4: parafraza
    5: spełnienie
    6: streszczenie
    7: tożsamość
    8: tło historyczne
    9: uszczegółowienie
    10: zawieranie
    11: źródło
\end{minted}
\small\begin{minted}[breaklines]{yaml}
5f34a87c-760e-4839-8ba8-7e643de1d607:
  instruction: 'Relacja między zdaniami.'
  demonstration: 'Zdanie_1: {sentence_1} Zdanie_2: {sentence_2} Etykieta: {labels}'
  label_mapping:
    0: brak relacji
    1: dalsze informacje
    2: krzyżowanie się
    3: opis
    4: parafraza
    5: spełnienie
    6: streszczenie
    7: tożsamość
    8: tło historyczne
    9: uszczegółowienie
    10: zawieranie
    11: źródło
\end{minted}
\small\begin{minted}[breaklines]{yaml}
a668bb22-b3c3-482c-809a-3f14eccdec1f:
  instruction: 'Która etykieta opisuje relację między zdaniami.

    Etykiety:

    - brak relacji

    - dalsze informacje

    - krzyżowanie się

    - opis

    - parafraza

    - spełnienie

    - streszczenie

    - tożsamość

    - tło historyczne

    - uszczegółowienie

    - zawieranie

    - źródło'
  demonstration: 'Zdania: {sentence_1}; {sentence_2} Etykieta: {labels}'
  label_mapping:
    0: brak relacji
    1: dalsze informacje
    2: krzyżowanie się
    3: opis
    4: parafraza
    5: spełnienie
    6: streszczenie
    7: tożsamość
    8: tło historyczne
    9: uszczegółowienie
    10: zawieranie
    11: źródło
\end{minted}
\small\begin{minted}[breaklines]{yaml}
debde233-ea35-4e23-9d2d-f49b2884f5b5:
  instruction: ''
  demonstration: 'Zdanie "{sentence_1}" ma się do zdania "{sentence_2}" jak {labels}'
  label_mapping:
    0: brak relacji
    1: dalsze informacje
    2: krzyżowanie się
    3: opis
    4: parafraza
    5: spełnienie
    6: streszczenie
    7: tożsamość
    8: tło historyczne
    9: uszczegółowienie
    10: zawieranie
    11: źródło
\end{minted}
\small\begin{minted}[breaklines]{yaml}
2ced9c63-7221-48b6-8c72-a8a78bbd13a0:
  instruction: 'Jakiego typu relacja występuje między parą zdań? Wybierz spośród:
    brak relacji, dalsze informacje, krzyżowanie się, opis, parafraza,
    spełnienie, streszczenie, tożsamość, tło historyczne, uszczegółowienie, zawieranie,
    źródło.'
  demonstration: '"{sentence_1}, "{sentence_2}, {labels}"'
  label_mapping:
    0: brak relacji
    1: dalsze informacje
    2: krzyżowanie się
    3: opis
    4: parafraza
    5: spełnienie
    6: streszczenie
    7: tożsamość
    8: tło historyczne
    9: uszczegółowienie
    10: zawieranie
    11: źródło
\end{minted}